\documentclass[11pt]{article}

\usepackage[preprint]{acl}

\usepackage{times}
\usepackage{latexsym}
\usepackage{booktabs}
\usepackage{colortbl}
\usepackage{orcidlink}
\usepackage{multirow}
\usepackage{soul}
\usepackage{amsmath}

\usepackage[most]{tcolorbox}
\usepackage{enumitem}
\usepackage{xcolor}
\usepackage[T1]{fontenc}
\usepackage{seqsplit}

\usepackage{listings}

\lstdefinestyle{jsonstyle}{
    basicstyle=\ttfamily\scriptsize,
    breaklines=true,
    breakatwhitespace=false,
    columns=fullflexible,
    keepspaces=true,
    showstringspaces=false,
    frame=single,
    framerule=0.3pt,
    rulecolor=\color{gray!50},
    backgroundcolor=\color{gray!4},
    xleftmargin=0.5em,
    xrightmargin=0.5em,
    aboveskip=0.6em,
    belowskip=0.6em,
    tabsize=2
}

\usepackage{algorithm}
\usepackage{algpseudocode}
\usepackage{float}

\usepackage{caption}

\usepackage[T1]{fontenc}

\usepackage[utf8]{inputenc}

\usepackage{microtype}

\usepackage{inconsolata}
\usepackage{graphicx}

\title{OVIBench: Benchmarking Online Video Question Answering under Interruption}

\author{
Naiming Liu$^{*1}$ \quad
Zhiheng Wu$^{*\S2}$ \quad
Shuning Wang$^{3}$ \quad
Tie Zhang$^{4}$
\\
\textbf{Bowen Liu$^{5}$ \quad
Tong Wang$^{\dagger2}$}
\\[2mm]
$^{1}$HIT,
$^{2}$CASIA,
$^{3}$ZJU,
$^{4}$UESTC,
$^{5}$HKUST \\
{\small
$^{*}$Equal contribution.
$^{\S}$Project leader.
$^{\dagger}$Corresponding author.
}
}

\begin{document}
\maketitle
\begin{abstract}
Recent vision language models (VLMs) have achieved strong progress in video understanding. However, most existing video QA research and benchmarks still follow an offline, single-round paradigm, overlooking realistic interactions where users may interrupt the model during answer generation. 
To address this gap, we formulate the task of Online Video Question Answering under Interruption and introduce \textbf{OVIBench}, the first standardized benchmark for evaluating VLMs in this setting.
OVIBench categorizes interruptions into three types: \textbf{Cancellation, False Trigger, Correction} and supports both \textbf{open-ended} and \textbf{multiple-choice} evaluations. To enable large-scale and reproducible testing, we develop an offline simulation protocol that reproduces interruption during generation under a unified temporal setup, together with a multi-dimensional metric suite for assessing interruption understanding and response generation. 
Experiments demonstrate that OVIBench effectively distinguishes models’ interruption-handling abilities, especially in following correction requests.
Finally, we construct a train set \textbf{OVI-Train} for interruption-aware fine-tuning. Models fine-tuned on this dataset achieve significant gains on OVIBench, validating the effectiveness of our benchmark and data design.
OVIBench, OVI-Train, and the evaluation code will be released.
\end{abstract}

\section{Introduction}
\label{sec:intro}

In recent years, multimodal language large models (MLLMs) have made significant progress in video understanding and video question answering~\cite{zhang2024simple, cai2024empowering}, enabling a range of applications such as live-stream interactions~\cite{yang2025visionzip, zhao2025accelerating}, online education~\cite{ray2025eduvidqa, ashutosh2024detours, nagarajan2024step}, and intelligent assistants~\cite{shu2025video, yang2025visionzip}. However, most existing research and evaluation benchmarks adopt an offline, single-round question-and-answer paradigm~\cite{zhou2025mlvu, shu2025video, lin2024streamingbench, niu2025ovo}. In this setting, the model observes the complete video $V$ and question $Q$ and generates the final answer $A$ in a single pass. This differs substantially from real online interactions~\cite{ma2025language, zhang2025omniflatten}, where users may issue new instructions or interruption signals while the model is generating answers, such as requests to stop or correct the ongoing response. In such cases, the model must not only continue generation but also infer the interruption intent and adjust its response strategy accordingly. The lack of a systematic evaluation framework for this dynamic setting makes it difficult to objectively characterize the capabilities of current MLLMs in interruptible interactions.

\begin{figure}[t]
\centering
\includegraphics[width=\linewidth]{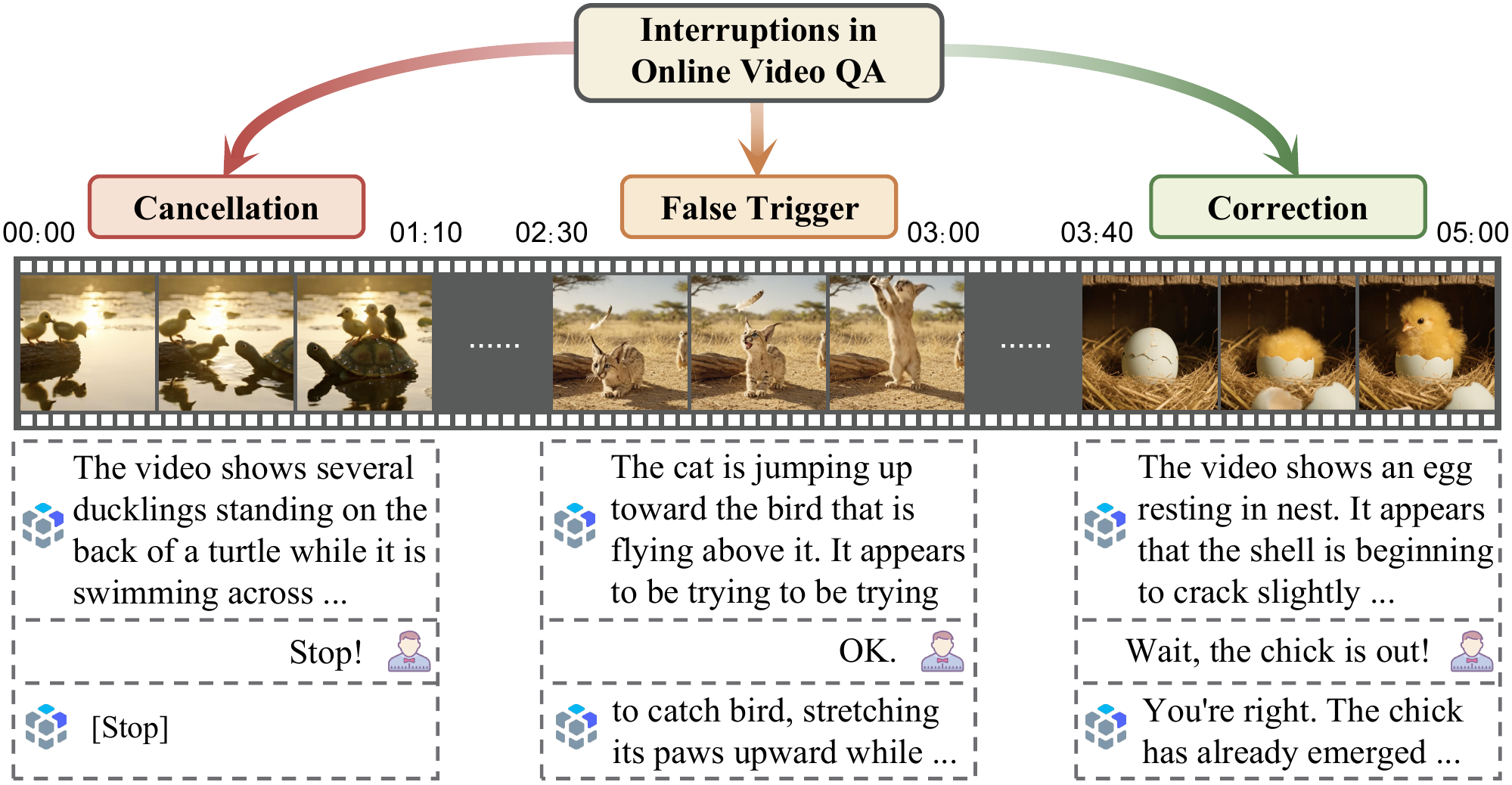}
\caption{
Illustration of three interruption types in online video question answering: Cancellation, False Trigger, and Correction.
In Online interactions, users may terminate the response, provide irrelevant signals, or introduce semantic corrections during generation.
}
\vspace{-2em}
\label{fig:interruption_taxonomy}
\end{figure}

To address this gap, we introduce \textbf{OVIBench}, the first standardized benchmark for online streaming video QA with interruptions. We formalize the task setting and categorize interruptions into three types: \textbf{Cancellation}, \textbf{False Trigger}, and \textbf{Correction} (Fig.~\ref{fig:interruption_taxonomy}). To evaluate interruption handling from both realistic and controllable perspectives, we build a complete data construction pipeline supporting two evaluation formats: \textbf{open-ended} and \textbf{multiple-choice}. The open-ended setting better reflects free-form real interactions, while the multiple-choice setting enables low-noise and reproducible evaluation for stable comparisons. Moreover, we develop an offline simulation protocol that reproduces ``interruption during generation'' under a unified temporal setup, enabling large-scale batch evaluation with high reproducibility.

For evaluation, we design a multi-dimensional metric suite spanning two aspects: \textbf{interruption understanding} and \textbf{response generation}. It measures key capabilities, including interruption-intent recognition, post-interruption intent fulfillment, pre/post-interruption fluency and content consistency, and consistency between the response and the observed video evidence. Experiments on OVIBench show that, while mainstream MLLMs perform well in standard offline video QA, they exhibit substantial weaknesses under online interruptions, particularly in reliably following user correction requests.

Furthermore, to investigate the root causes and explore effective improvements, we construct \textbf{OVI-Train}, a multiple-choice training dataset for interruption-aware fine-tuning, and apply targeted fine-tuning to a base model. Compared with Qwen2.5-VL-7B, our 7B model achieves substantial gains, improving multiple-choice accuracy by 16.88\% and interruption-type accuracy by 13.94\%, and even surpassing the 72B model in the same series. These results validate our data and task design and provide a reusable foundation for future training and evaluation. As the first benchmark dedicated to interruption scenarios in online video question answering, OVIBench pushes large video models beyond offline, static inference toward more realistic online interaction. The following contributions are made in this paper.
\begin{itemize}
    \item We introduce \textbf{OVIBench}, the first benchmark formalizing interruptions in online video QA, supporting both open-ended and multiple-choice evaluation.
    \item We propose an \textbf{offline simulation} protocol for interruptions to enable large-scale, low-noise evaluation, and design a \textbf{multi-dimensional metric suite} to comprehensively assess model behavior under dynamic interactions.
    \item We present \textbf{OVI-Train}, a multiple-choice training dataset for interruption scenarios. Experiments validate the effectiveness of our data and task design.
\end{itemize}

\section{Related Work}
\label{sec:related}

\noindent\textbf{Video-LLMs and Online Video Understanding.}
Recently, the capabilities of MLLMs have been extended from static images to action reasoning and cross-modal modeling in complex video scenes~\cite{ren2024timechat, li2024llama, qian2024streaming}. 
This progress has been driven by more sophisticated visual perception modules and staged training strategies, which have substantially improved models' offline video understanding capabilities~\cite{wang2024internvideo2, maaz2024video, team2024gemini,liu2023llava}. For example, Video-LLaVA~\cite{lin2024video} focuses on cross-modal spatio-temporal modeling, Qwen-VL~\cite{wang2024qwen2} performs well in multi-modal modeling, GPT-4o~\cite{hurst2024gpt} demonstrates strong long-form text generation capability, and VideoChat~\cite{li2025videochat} and VideoLLaMA~\cite{zhang2025videollama} achieve strong performance in spatio-temporal reasoning. However, in mainstream frameworks, video understanding is usually based on complete observation-holistic generation offline reasoning, where the model generates a single response after receiving the complete video~\cite{wang2024internvideo2}. This limits the model's ability to cope with dynamic interruptions and context updates in real interaction environments.

\noindent\textbf{Video Benchmarks.}
Existing video understanding benchmarks can be broadly categorized into offline and online evaluation settings. Video-MME~\cite{fu2025video} evaluates the comprehensive video understanding ability of multimodal models through multiple-choice questions across diverse domains and durations. LVBench~\cite{wang2025lvbench} focuses on long-video reasoning and assesses whether models can retrieve relevant evidence from extended video contexts. More recently, online benchmarks moved toward streaming inputs and multi-round interactions. StreamingBench~\cite{lin2024streamingbench} evaluates models under streaming video inputs and measures their ability to maintain consistent reasoning as new frames arrive. OVO-Bench~\cite{niu2025ovo} studies online video understanding with progressive observations and emphasized adaptive reasoning under evolving visual contexts. OVBench~\cite{huang2025online} further introduces interaction-oriented evaluation protocols to assess model behaviors in online video question answering scenarios. Although these benchmarks covered multiple aspects of video understanding, existing evaluation frameworks ignore the issue of interruption handling in dynamic interactions.

\noindent\textbf{Interactive Modeling and Turn-Taking Benchmarks.}
In the research of human-computer speech dialogue, turn-taking is usually regarded as the core mechanism of natural interaction~\cite{lin2022duplex,lin2025full, arora2025talking, zheng2023judging}. Recent benchmarks such as Talking Turns builds a conversation event prediction and evaluation framework for audio basic models~\cite{arora2025talking}. Full-Duplex-Bench systematically evaluates full-duplex conversation models from multiple dimensions such as pause processing, backchannel behavior, and interrupt response~\cite{lin2025full}. AudioBench extends to a wider range of speech and paralinguistic ability assessments~\cite{wang2025audiobench}. These works emphasize the temporal behavioral evaluation of full-duplex speech dialogue models, but their research objects mainly focus on audio or text modalities. In the video stream question and answer scenario, interruption is not only a problem of judging the timing of speaking, but also a multi-modal decision-making problem based on visual context. So far, there is no benchmark to evaluate this scenario under a unified framework.

\section{OVIBench}

\subsection{Task Formulation}

We define Online Video Question Answering under Interruption to evaluate how a model handles user interruptions while answering questions about online video.
Online video stream is denoted as $V = \{v_1, v_2, \dots, v_T\}$, where $v_t$ represents the video frame at time step $t$. At time $t_q$, a user issues an initial question $Q_0$, upon which the model begins generating an answer in an autoregressive manner, denoted as $A = \{a_1, a_2, \dots\}$.
During generation, at time $t_i > t_q$, the user may provide an interruption signal $I$. At this moment, the model has produced a partial response $A_{\le t_i} = \{a_1, \dots, a_{t_i}\}$. Therefore, the observable context at interruption time is the video frames $V_{\le t_i}$, the initial question $Q_0$, the partially generated answer $A_{\le t_i}$, and the interruption signal $I$.
Under this setting, the model is required to perform intent-response coupled modeling. Specifically, it first identify interrupt type $c \in \mathcal{C}$, where $\mathcal{C} = \{\text{False Trigger}, \text{Correction}, \text{Cancellation}\}$.
Conditioned on the observable context and interrupt type $c$, the model then executes the corresponding strategy (e.g., continue, revise, or terminate), producing the post-interruption response $A' = f(V_{\le t_i}, Q_0, A_{\le t_i}, I, c)$. 
 Unlike conventional offline VideoQA~\cite{wu2024longvideobench, chen2024videollm}, which can be formulated as a one-shot mapping $A = f(V, Q)$, OVI introduces answer-time external intervention. The model must handle partially generated answers and an evolving interaction state. This makes the task online and state-aware, rather than a static one-shot inference problem.

\subsection{Interruption Taxonomy}

In online video question answering, interruptions are not merely temporal disturbances (e.g., meaningless insertions), but semantic inputs that may alter the ongoing question-answering process.
To characterize model behavior under different interruption conditions, we construct a structured interruption taxonomy. This taxonomy is defined based on two criteria: (1) whether the interruption changes the current semantic context, i.e., whether it introduces new facts, constraints, or objectives; and (2) whether it requires the model to adjust its generation strategy, such as continuing the original response, revising existing content, or terminating the response process.
Formally, we define an interruption as an external input occurring during response generation, whose role is to impose constraints on or modify the current interaction state. Based on the degree to which an interruption affects the contextual state and strategy space, we categorize interruptions into three types:

\noindent
\textbf{False Trigger.} This type does not change the semantic context or the generation objective. The model is expected to recognize the interruption as ineffective and continue the original response. This type evaluates the model‘s ability to resist spurious interventions and maintain generation stability.

\noindent
\textbf{Correction.} This type introduces new semantic constraints that modify previously generated content or the question context. The model must update its interaction state and revise the response coherently under the new conditions. This type assesses contextual reconstruction and response-consistent revision.

\noindent
\textbf{Cancellation.} This type explicitly terminates the current query and invalidates the ongoing response. The model should stop or reset generation upon recognition. This type evaluates responsiveness at the interaction-control level.

\subsection{Data Collection}

\begin{figure}[t]
\centering
\includegraphics[width=0.85\linewidth]{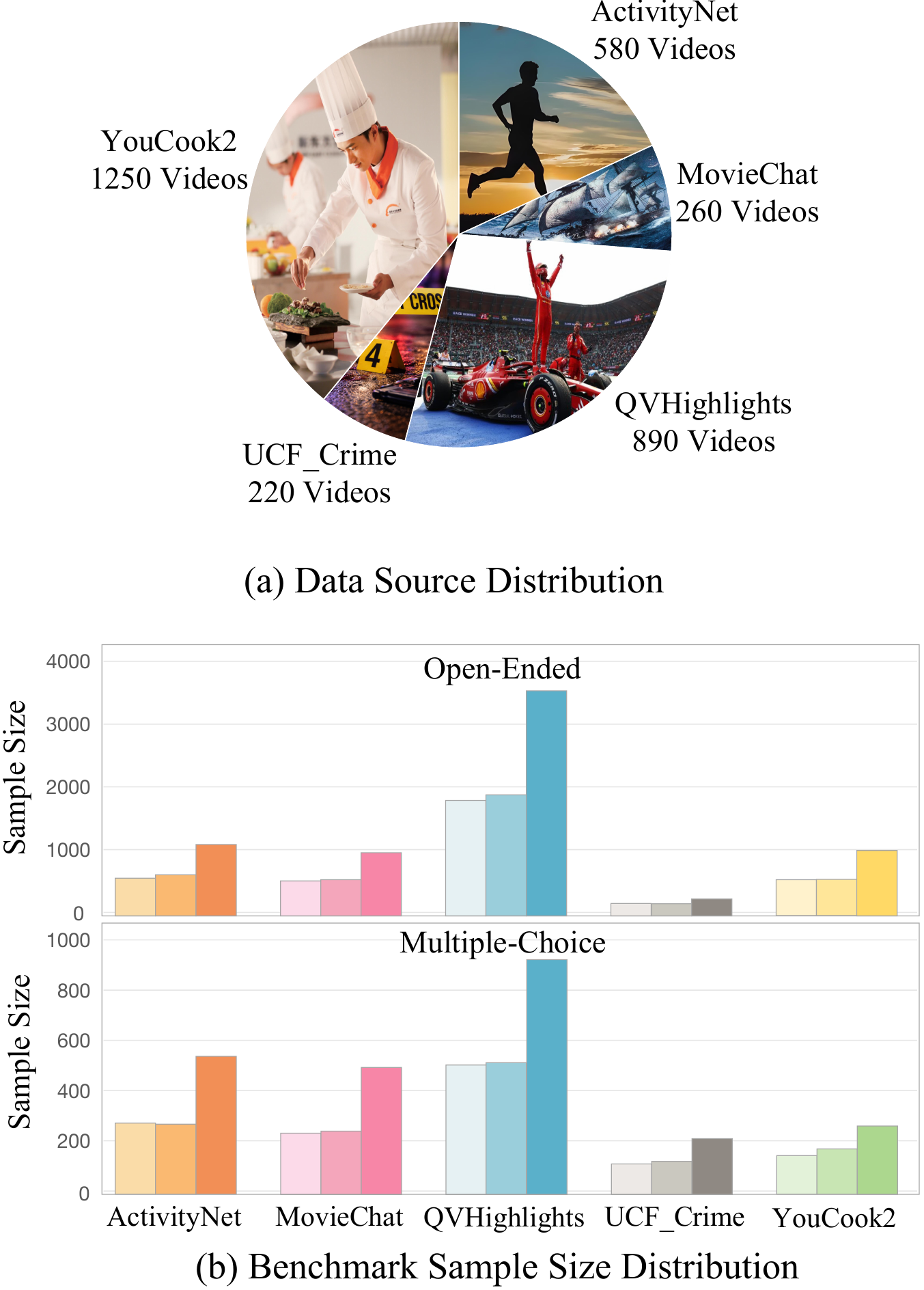}
\caption{
(a) Distribution of video sources across five public datasets during data collection.
(b) Sample size distribution of the benchmark under open-ended and multiple-choice evaluation settings. For each dataset, three bars are shown from left to right, corresponding to the False Trigger, Cancellation, and Correction interruption types.
}
\label{fig:data_distribution}
\end{figure}

To construct an online video QA dataset with interruptions during generation, we curate 3,200 videos from five public datasets:
ActivityNet~\cite{caba2015activitynet}, MovieChat~\cite{song2024moviechat},
QVHighlights~\cite{lei2021detecting}, UCF-Crime~\cite{qian2025ucf},
and YouCook2~\cite{ohkawa2025exo2egodvc}. 
These sources cover diverse video understanding scenarios, including open-domain activities, narrative dialogue, highlight localization, anomalous events, and instructional procedures. For each source dataset, videos are randomly sampled without replacement to ensure unbiased coverage of content diversity. We apply basic quality filtering before inclusion: videos with corrupted files, extremely short duration (less than 10 seconds), missing visual content, or severe frame loss are excluded. This collection provides diverse temporal and semantic contexts for constructing False Trigger, Correction, and Cancellation interruptions. The detailed source distribution is shown in Fig.~\ref{fig:data_distribution}(a).

\subsection{QA Generation and Data format}
\label{sec:data_generation}

\begin{figure*}[t]
\centering
\includegraphics[width=0.95\linewidth]{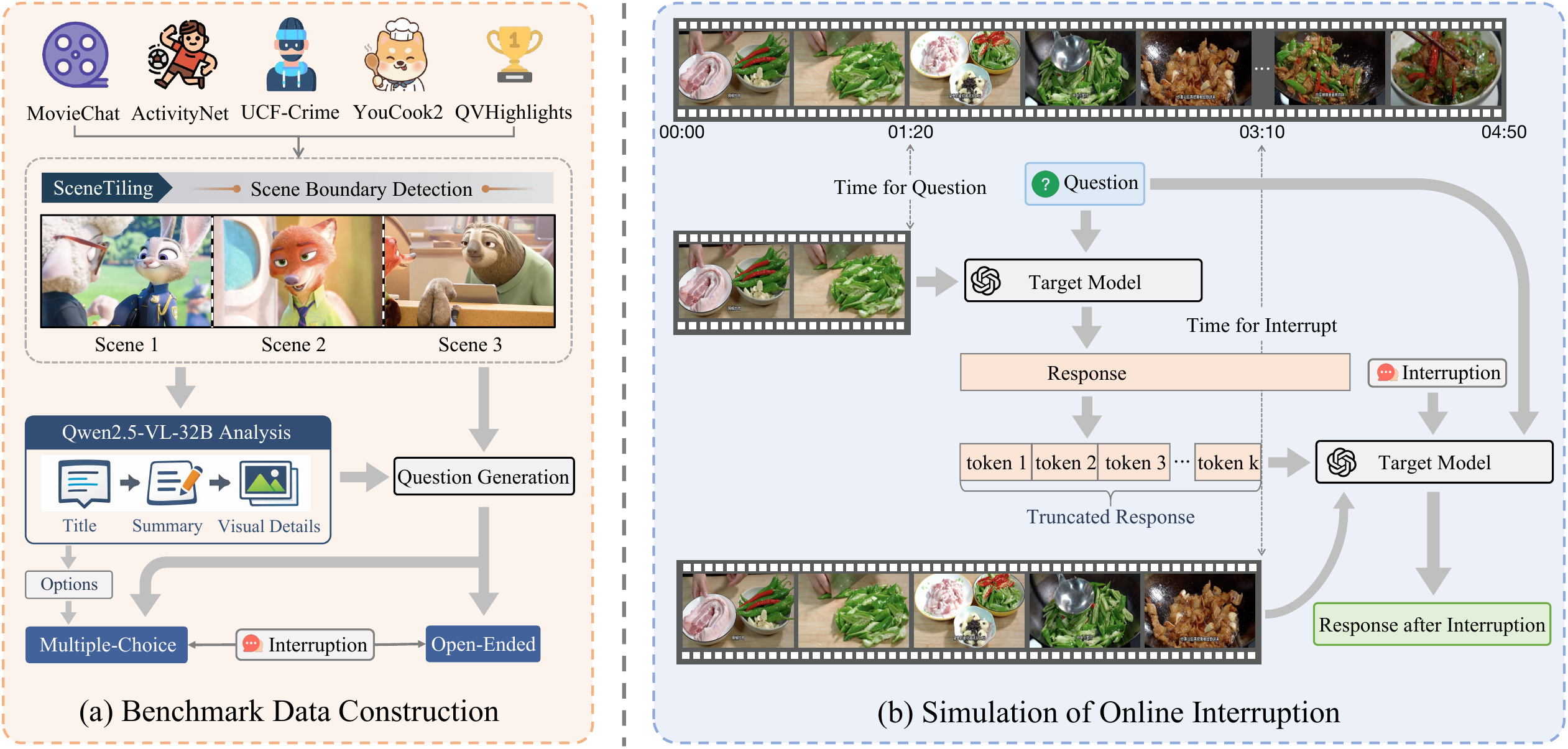}
\caption{
Overview of OVIBench. 
(a) Benchmark data construction pipeline, including scene segmentation,
VLM-based question generation and interruption insertion for both
open-ended and multiple-choice formats.
(b) Offline simulation of online video interruption, where the model generates responses autoregressively and reacts to interruption signals during generation.
}
\label{fig:ovibench_pipeline}
\end{figure*}

As shown in Fig.~\ref{fig:ovibench_pipeline}(a), we first use a vision language model (VLM) to generate a complex question $Q$ for each video sample $V$. By increasing the complexity of the question, we ensure that the model can generate a raw answer $A$ of sufficient length, thus reserving enough processing space for the interruption insert. The video timestamp at which the question is triggered is denoted as $T_0$. Next, based on the preset interruption type, we use the VLM to generate the corresponding interruption signal $I$. The interruption time $T_1$ is randomly generated within the interval $(T_0, T_{end})$ ($T_{end}$ is the video end time). Finally, each open-ended sample is encapsulated in a standard format: $Data_{open} = \{Q, T_0, I, T_1, V\}$.

In addition, we construct an evaluation set in a multiple-choice format. Based on the data format described above, we introduce candidate options $O$ and the ground truth label $GT$. The data format is $Data_{choice}=\{Q, T_0, I, T_1, V, O, GT\}$. Options $O$ contain four categories: three categories are correct answers generated by three different interruption types  (Cancellation, False Trigger, and Correction), and the remaining one is an incorrect distractor generated by a VLM. The ground truth $GT$ is the candidate that matches the inserted interruption type.
To improve robustness, we randomly shuffle the option order for all questions.

Ultimately, we use Qwen2.5-VL-32B~\cite{yao2024fine} to generate 13145 samples for $Data_{\text{open}}$ and 4710 samples for $Data_{\text{choice}}$.
Since the open-ended format better reflects real-world usage, $Data_{\text{open}}$ more closely matches practical interrupted interactions and is scored by a judge model.
In contrast, the multiple-choice set reduces evaluation noise and bias, enabling more accurate and fair comparisons across models.
The distributions of the two evaluation sets are shown in Fig.~\ref{fig:data_distribution}(b).
Details and examples of the data generation process are provided in the appendix.

\label{sec:ovibench}

\subsection{Simulate Online Video Interruption}

Since offline batch evaluation is not feasible in real-world dynamic interactions, we simulate the online video interruption process offline. As shown in Fig.~\ref{fig:ovibench_pipeline} (b), the simulation process is as follows:

First, we determine the video content that the model can observe based on the timestamp $T_0$ of question $Q$ in the data. This step simulates the video stream transmission process, ensuring that the model receives the video segment $V_Q$ that is visible at the given time.

Then, we input the observable video $V_Q$ along with question $Q$ into the model to generate the answer $A$ for question $Q$. At this point, the model infers based on the currently visible video content and generates a preliminary answer.

Next, we simulate the interruption process. We set a fixed generation rate for answer $A$, which allows us to calculate the time at which each token is generated and simulate the time progression of answer $A$ during its generation. Using the interruption timestamp $T_1$ in the data, we determine the exact time point when the interruption occurs. With $T_1$, we infer the video content $V_I$ that the model has already observed at this moment, and the position where the generated answer $A$ was truncated.

Finally, we input the interruption signal $I$, the observable video $V_I$, and the truncated answer $A_I$ into the model. Based on this information, the model determines the intent of the interruption signal (e.g., whether it's a false trigger, cancellation, or correction) and generates a new response. The model reacts to the interruption, and we evaluate the model's response, measuring its performance and adaptability in handling interruptions.

Through this offline simulation, we can assess the model's ability to answer questions, its responsiveness to interruption signals, and its adaptability under various interruption scenarios.

\section{Model-Judged Evaluation of OVIBench}
\label{sec:open_eval}

\subsection{Comparison Methods}
To evaluate the performance of existing models in the online video question answering under interruption scenario, we select several state-of-the-art VLMs for comparison, including Doubao-Seed-1.6~\cite{huang2025memorb}, Gemini-2.5-Flash~\cite{comanici2025gemini}, Qwen2.5-VL-7B~\cite{yao2024fine}, Qwen3-VL-8B~\cite{bai2025qwen3}, VideoChat-R1-7B~\cite{li2025videochat}, VideoLLaMA3-7B~\cite{zhang2025videollama}, and Qwen3-VL-30B-A3B~\cite{bai2025qwen3}. These models represent different architectures and application domains of current vision-language models. By comparing them, we can comprehensively assess the strengths and weaknesses of existing models in interruption scenarios and reveal the performance gaps in real-time online video question answering.

\subsection{Metrics Design}
\label{sec:metrics}

In online video QA, interrupt signals may occur at any time during answer generation. Handling such interruptions requires models not only to produce correct content but also to recognize user intent. 
In addition, the model needs to adjust its response logic to be consistent with visual and contextual information. The above capabilities cannot be measured by a single indicator. Therefore, beyond intent recognition accuracy, we introduce six complementary metrics to assess the model from multiple perspectives. All metrics are automatically scored by a judge model based on predefined criteria.

\noindent \textbf{Intent-Action Consistency (IAC).}
IAC evaluates whether the model's post-interruption response strategy aligns with the interruption intent it has identified in online video interaction.
For example, if the model interprets the user‘s intent as Cancellation but continues generating content or instead performs a correction, the response is considered inconsistent.

\noindent \textbf{Textual Fluency Score (TFS).}
TFS is used to evaluate whether the model‘s post-interruption responses can smoothly continue from the pre-interruption responses at the textual dimension.

\noindent \textbf{Visual Evidence Score (VES).}
VES is used to evaluate whether the visual content mentioned in the model's post-interruption response is consistent with the observed video evidence.

\noindent \textbf{Intent Fulfillment Accuracy (IFA).}
IFA is used to evaluate whether the model‘s post-interruption response fulfills the user‘s intent. Even if the model misidentifies the intent, the response is still considered correct as long as it successfully satisfies the user‘s actual intent.

\noindent \textbf{Correction Following Score (CFS).}
CFS measures the extent to which, after a user issues a correction-type interruption, the model‘s post-interruption response incorporates and executes the requested corrections. For example, replacing incorrect information, filling in missing elements, or removing content that should not appear.

\noindent \textbf{Contextual Consistency Score (CCS).}
CCS measures whether the post-interruption responses generated by the model are consistent with the existing context at the content dimension.

\subsection{Judge Model}
\label{sec:judge}

\begin{table*}[ht]
\centering
\small
\caption{MAE between different Judge models and human annotations on the validation set, and lower is better. The best value in each column is highlighted in bold.}
\setlength{\tabcolsep}{5pt} 
\renewcommand{\arraystretch}{1.4} 
\label{tab:judge_selection}
\resizebox{0.95\linewidth}{!}{
\begin{tabular}{lcccccc}
\toprule
\textbf{Model} & \textbf{MAE$_{\text{IAC}}$} & \textbf{MAE$_{\text{TFS}}$} & \textbf{MAE$_{\text{IFA}}$ } & \textbf{MAE$_{\text{CCS}}$ } & \textbf{MAE$_{\text{CFS}}$ } & \textbf{MAE$_{\text{VES}}$ } \\
\midrule
Qwen3-VL-8B~\cite{bai2025qwen3}      & 0.2070 & 1.1445 & 0.1641 & 3.1289 & 2.0703 & 1.6016 \\
Qwen3-VL-30B-A3B~\cite{bai2025qwen3} & 0.1406 & 0.8984 & 0.1250 & 3.1602 & 2.3359 & \textbf{0.8477} \\
Qwen3-VL-235B-A22B~\cite{bai2025qwen3}        & \textbf{0.0508} & \textbf{0.3672} & \textbf{0.0625} & \textbf{0.9453} & \textbf{0.5547} & 1.0273 \\
Doubao-Seed-1.6~\cite{huang2025memorb}  & 0.0859 & 1.0859 & 0.1484 & 2.6016 & 1.9766 & 2.1445 \\
Gemini-2.5-Flash~\cite{comanici2025gemini} & 0.0938 & 1.2656 & 0.1406 & 2.8984 & 1.8711 & - \\
GPT-4.1-mini~\cite{achiam2023gpt}  & 0.0938 & 0.7500 & 0.1445 & 3.0938 & 1.6484 & - \\
\bottomrule
\end{tabular}
}
\end{table*}

In the open-ended benchmark, we use a model-judged approach to automatically score the model's response~\cite{liu2023g, chehbouni2025neither}. For each metric, we designed a corresponding Judge Prompt (See appendix for details). IAC and IFA are computed using a binary satisfaction criterion (0/1), while TFS, VES, CFS, and CCS use a 1–10 scoring, with higher values indicating better performance.
To select a stable Judge model with higher consistency with human reviewers, 415 samples are selected from the open-ended subset of OVIbench to construct a validation set. Qwen3-VL-8B, Qwen3-VL-30B-A3B, Qwen3-VL-235B-A22B~\cite{bai2025qwen3}, Doubao-Seed-1.6, Gemini-2.5-Flash, and GPT-4.1-mini~\cite{achiam2023gpt} are evaluated as Judge model. We measure judge performance using the mean absolute error (MAE) between the judge scores and human annotations (Table~\ref{tab:judge_selection}).
Experiments reveal that Qwen3-VL-235B-A22B demonstrated the best consistency with human reviewers for metrics IAC, TFS, IFA, CFS, and CCS; while Qwen3-VL-30B-A3B performed best for VES. Therefore, we ultimately adopt Qwen3-VL-235B-A22B as the Judge Model for all metrics except VES, and specifically use Qwen3-VL-30B-A3B as the Judge Model for VES.

\subsection{Results and Analysis}
\label{sec:open_results}

Table~\ref{tab:main_results} reports the results on the open-ended subset of OVIBench. Overall, existing models show limited capability in handling interruptions in online video question answering, and their performance varies substantially across metrics. Most models obtain low scores on CFS and CCS, indicating persistent challenges in incorporating and executing user-requested corrections and maintaining content-level consistency with the existing context.
Doubao-Seed-1.6 achieves the best overall results, ranking highest in DAC, TFS, CFS, CCS, and overall accuracy.  
Qwen3-VL-8B achieves the best VES and IFA, suggesting strong video–response consistency and effective intent fulfillment.
However, all models still exhibit substantial limitations in following user correction instructions and revising their responses accordingly.
These results indicate that interruption handling remains challenging for current MLLMs.

\begin{table*}[ht]
\centering
\caption{Experimental results on the open-ended subset of OVIBench.
CA represents the accuracy of interruption-type classification.
IAC and IFA are binary classification accuracy. TFS, VES, CFS, and CCS are scored on a 1-10 scale.
Higher values indicate better performance ($\uparrow$).
The best results in each column are highlighted in bold.}
\label{tab:main_results}

\setlength{\tabcolsep}{2pt}
\renewcommand{\arraystretch}{1.4}

\resizebox{0.8\textwidth}{!}{
\begin{tabular}{l*{7}{c}} 
\toprule
\rowcolor[HTML]{F9F9F9} \textbf{Model/Metric} & \textbf{IAC $\uparrow$} & \textbf{TFS $\uparrow$} & \textbf{VES $\uparrow$} & \textbf{IFA $\uparrow$} & \textbf{CFS $\uparrow$} & \textbf{CCS $\uparrow$} & \textbf{CA $\uparrow$} \\
\midrule
Doubao-Seed-1.6~\cite{huang2025memorb} & \textbf{95.73\%} & \textbf{9.61} & 6.23 & 79.42\% & \textbf{5.24} & \textbf{7.26} & \textbf{92.85}\% \\
\rowcolor[HTML]{F2F2F2} 
Gemini-2.5-Flash~\cite{comanici2025gemini} & 76.45\% & 8.41 & 6.28 & 64.58\% & 4.42 & 6.97 & 73.39\%  \\
Qwen2.5-VL-7B~\cite{yao2024fine} & 64.14\% & 7.55 & 5.34 & 50.11\% & 3.73 & 5.06 & 83.64\%  \\
\rowcolor[HTML]{F2F2F2} 
Qwen3-VL-8B~\cite{bai2025qwen3} & 93.63\% & 9.32 & \textbf{6.68} & \textbf{81.07\%} & 4.91 & 6.75 & 88.61\%  \\
VideoChat-R1-7B~\cite{li2025videochat} & 66.42\% & 7.69 & 5.36 & 51.03\% & 3.86 & 4.98 & 87.27\%  \\
\rowcolor[HTML]{F2F2F2} 
VideoLLaMA3-7B~\cite{zhang2025videollama} & 72.33\% & 7.50 & 5.25 & 57.92\% & 4.06 & 5.63 & 89.61\%  \\
Qwen3-VL-30B-A3B~\cite{bai2025qwen3} & 88.29\% & 9.05 & 6.50 & 80.23\% & 5.10 & 6.47 & 91.10\% \\
\bottomrule
\end{tabular}
}

\end{table*}

\begin{figure}[ht]
\centering
\includegraphics[width=0.9\linewidth]{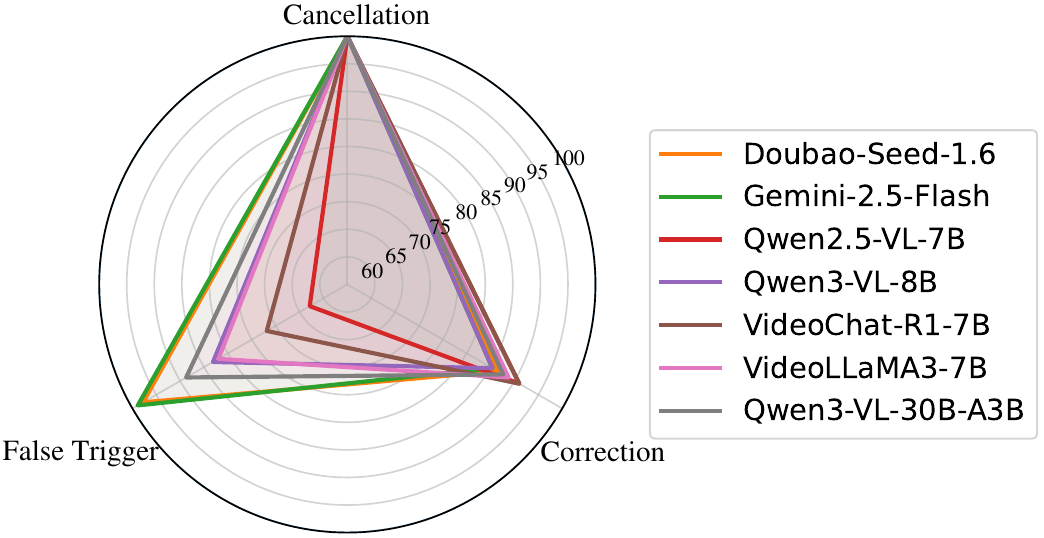}
\caption{
Comparison of model performance across different interruption types on OVIBench. Each axis represents the recognition accuracy (\%) for one interruption type, and larger areas indicate better overall interruption understanding capability.
}
\vspace{-1em}
\label{fig:acc}
\end{figure}

Fig.~\ref{fig:acc} shows the interruption-type classification accuracy (CA) for different interruption categories. 
For Cancellation, all models achieve nearly perfect accuracy, reaching close to 100\%. This result indicates that explicit termination signals are easy to recognize. 
In contrast, False Trigger exhibits the largest performance variation across models. Smaller models show noticeably lower accuracy, indicating that distinguishing irrelevant interruptions from meaningful ones remains challenging. Correction lies between Cancellation and False Trigger in difficulty. Most models achieve accuracy around 87\%–92\%, suggesting that corrective intent is recognizable but still more ambiguous than explicit termination signals.

\section{Multiple-Choice Evaluation of OVIBench}
\label{sec:closed_eval}

\subsection{Interruption-aware Fine-tuning}
\label{sec:OVI-Train}

\textbf{Training Dataset.} Overall, the suboptimal performance of MLLMs can be largely attributed to the lack of explicit interruption-type supervision in instruction-tuning data. To address this issue, we construct 20765 training instances (OVI-Train) following the data generation pipeline in Sec.~\ref{sec:data_generation}. Specifically, we first derive a target answer $A$ for each question $Q$ in the dataset, where an interruption is intended to occur during generation. Then, based on the offline simulation procedure in Sec. 3.5, we truncate $A$ at the simulated interruption point and inject an interruption instruction, yielding training samples of the form $Data_{train}=\{Q, A_I, I, V_I, O, GT\}$. Here, $A_I$ is the truncated answer segment shown before the interruption, $I$ is the interruption signal/instruction, $V_I$ denotes the video segment at the interruption moment, and $O$ is the set of candidate options. During training, we feed $\{Q, A_I, I, V_I, O\}$ as model inputs and use $GT$ as the supervision 
label, enabling the model to accurately recognize and select the correct interruption type in online video interactions. The data distribution is shown in Table~\ref{tab:OVI-Train_data_stats}.

\begin{table}[t]
\centering
\caption{Statistics of OVI-Train instances by video source and interruption type.}
\label{tab:OVI-Train_data_stats}
\resizebox{1\linewidth}{!}{

\begin{tabular}{lcccc}
\toprule

& Cancellaton 
& Correction 
& False Trigger 
& Total \\

\midrule
ActivityNet & 840 & 1616 & 799 & 3255 \\
MovieChat & 475 & 913 & 477 & 1865 \\
QVHighlights & 2736 & 5365 & 2744 & 10845 \\
UCF\_Crime & 173 & 297 & 140 & 610 \\
YouCook2 & 1041 & 2050 & 1099 & 4190 \\
\bottomrule
\end{tabular}
}
\end{table}

\begin{table*}[t]
\centering
\caption{Multiple-choice evaluation results on OVIBench. MA denotes answer selection accuracy, and CA denotes interruption-type classification accuracy. $\Delta$ indicates the performance improvement of our model over the baseline. The best results in each column are highlighted in bold.}
\label{tab:closed_set_results}

\resizebox{0.95\textwidth}{!}{
\renewcommand{\arraystretch}{1.3}
\begin{tabular}{lcccccccc}
\toprule
\textbf{Model} & \multicolumn{2}{c}{\textbf{All}} & \multicolumn{2}{c}{\textbf{False Trigger}} & \multicolumn{2}{c}{\textbf{Cancellation}} & \multicolumn{2}{c}{\textbf{Correction}} \\
\cline{2-9}
& \textbf{MA} & \textbf{CA} & \textbf{MA} & \textbf{CA} & \textbf{MA} & \textbf{CA} & \textbf{MA} & \textbf{CA} \\
\midrule
\rowcolor[HTML]{F0F0F0} 
\multicolumn{9}{c}{\textit{Open-Source Models}} \\
\midrule
Qwen2.5-VL-7B~\cite{yao2024fine} 
& 65.54 & 81.39 & 43.46 & 98.69 & 45.69 & 42.74 & 86.99 & 92.79 \\
Qwen2.5-VL-72B~\cite{yao2024fine} 
& 71.46 & 93.36 & 95.37 & \textbf{99.16} & 30.35 & \textbf{98.49} & 65.93 & 77.24 \\
VideoChat-R1-7B~\cite{li2025videochat} 
& 71.92 & 83.67 & 56.54 & 98.60 & 52.87 & 53.55 & 89.58 & 91.82 \\
Qwen3-VL-30B-A3B~\cite{bai2025qwen3} 
& 67.75 & 84.47 & 15.62 & 66.93 & 65.03 & 73.65 & 95.43 & 98.95 \\
\rowcolor[HTML]{F0F0F0} 
\midrule
\multicolumn{9}{c}{\textit{Closed-Source Models}} \\
\midrule
Doubao-Seed-1.6~\cite{huang2025memorb}  
& 66.34 & 93.01 & 30.28 & 98.25 & 43.50 & 77.62 & 96.40 & 98.37 \\
Gemini-2.5-Flash~\cite{comanici2025gemini} 
& 52.94 & 63.71 & 56.63 & 63.61 & 17.40 & 50.76 & 69.58 & 70.51 \\
\midrule
\textbf{Ours (7B)} 
& \textbf{82.42} & \textbf{95.33} 
& \textbf{96.40} & 98.81
& \textbf{77.24} &91.82
& \textbf{98.69} & \textbf{98.83} \\
\textbf{$\Delta$ (vs Qwen2.5-VL-7B)} 
& +16.88 & +13.94 
& +52.94 & +0.12 
& +31.55 & +49.08 
& +11.70 & +6.04 \\
\bottomrule
\end{tabular}
}
\end{table*}

\noindent \textbf{Training Details.}
We perform LoRA-based fine-tuning~\cite{dettmers2023qlora} on the Qwen2.5-VL-7B base model using the OVI-Train dataset.
The LoRA rank $r$ and scaling factor $\alpha$ are set to 16 and 32. The training process uses the AdamW optimizer with an initial learning rate of $1\times10^{-4}$, together with a cosine learning rate schedule and a 3\% linear warmup phase. The fine-tuning is conducted on 8 NVIDIA H20 GPUs with a global batch size of 128, trained for 10 epochs.

\subsection{Metrics}
To measure a model‘s interruption handling ability during generation under reproducible, low-noise conditions, we evaluate the multiple-choice subset of OVIBench using two core metrics. We report (1) Multiple-Choice Question Accuracy (MA) and (2) Interruption-Type Classification Accuracy (CA).
This multiple-choice setting complements the open-ended evaluation by reducing subjectivity via discrete options while explicitly measuring intent recognition under interruption. MA measures whether the model selects the correct answer option from the candidates under online video interruption. CA measures whether model correctly identifies the interruption intent (i.e., the interruption type).

\subsection{Results and Analysis}

To evaluate the effectiveness of the proposed fine-tuning under interruption scenarios, we conduct a multiple-choice evaluation on Qwen2.5-VL-7B. We compare it with several vision-language models, including Doubao-Seed-1.6, Gemini-2.5-Flash, Qwen2.5-VL-72B, Qwen3-VL-30B-A3B, and VideoChat-R1-7B. 

As shown in Table~\ref{tab:closed_set_results}, the Qwen2.5-VL-7B baseline shows limited performance, especially on MA under False Trigger and Cancellation. Its overall MA and CA reach only 65.54\% and 81.39\%, indicating limited capability in handling interruption scenarios. After fine-tuning, our 7B model achieves the best overall results, reaching 82.42\% MA and 95.33\% CA on the full set. The largest gains appear in False Trigger MA (+52.94\%) and Cancellation CA (+49.08\%). These improvements demonstrate the effectiveness of the OVI-Train dataset for interruption-aware training. Notably, the fine-tuned 7B model surpasses larger models such as Qwen2.5-VL-72B and also outperforms video-specialized models such as VideoChat-R1-7B on several metrics. This result indicates that targeted interruption-aware fine-tuning can significantly enhance online video question answering capability even with a smaller model.

\section{Conclusion}
This paper studies online video question answering, where users may interrupt the model during response generation. To analyze this scenario, we introduce the Online Video Question Answering under the Interruption task and construct the benchmark OVIBench, which models three interruption types and supports both open-ended and multiple-choice evaluation. We simulate interruptions offline to enable large-scale, reproducible evaluation, and design six complementary metrics to comprehensively assess model performance. Experiments show that existing MLLMs still struggle under interruptions. To address this, we introduce the OVI-Train dataset for interruption-aware fine-tuning and demonstrate that specialized training significantly improves model responsiveness. Overall, this study provides a practical framework to evaluate and improve video QA models in real-world scenarios where user interruptions may occur.

\section{Limitations}
Although OVIBench provides a systematic and reproducible framework for evaluating interruption handling in online video question answering, several directions remain for further extension. We simulate user interruptions during model generation through an offline protocol, which enables fair comparison across models, but real online systems may involve more dynamic generation speed, network latency, and user input timing. In addition, this work focuses on three representative interruption types, namely Cancellation, False Trigger, and Correction, while more complex scenarios such as multi-turn follow-up questions, task switching, and mixed intentions can be explored in future work. Moreover, OVIBench relies on VLMs to generate questions, interruption signals, and candidate answers, which supports scalable and consistent data construction, while incorporating more human annotations or real user interaction data could further improve its naturalness and coverage. Finally, although we evaluate multiple mainstream VLMs and validate the effectiveness of OVI-Train, its generalization to broader model architectures, languages, and real-world interactive settings involving speech or multi-user inputs remains an important direction for future research.


%
%
\bibliography{custom}

\clearpage
\appendix

\section{Prompt Templates}
Intent-Action Consistency asks the judge model to determine whether the model’s post-interruption reaction is logically consistent with its predicted interruption type. All placeholders are filled with sample-specific information during inference.

\begin{tcolorbox}[
    colback=gray!4,
    colframe=gray!75,
    title={IAC Evaluation Prompt},
    fonttitle=\bfseries,
    boxrule=0.5pt,
    arc=1.5mm,
    left=2mm,right=2mm,top=1mm,bottom=1mm,
    toptitle=0.5mm,
    bottomtitle=0.5mm,
    before skip=7mm,
    after skip=7mm
]
\small

\textbf{Role.} 
You are a professional dialogue system interaction evaluation expert. You are assessing a vision-language model’s \textbf{Intent-Action Consistency} in real-time video stream interaction.

\vspace{0.4em}
\textbf{Interaction Background.}\\
Original question: \texttt{<question>} (timestamp: \texttt{<q\_ts>}s)\\
Response before interruption: \texttt{<truncated\_text>}\\
User interruption signal: \texttt{<interruption\_signal>} (timestamp: \texttt{<i\_ts>}s)

\vspace{0.4em}
\textbf{Model Behavior.}\\
Predicted interruption type: \texttt{<pred\_type>}\\
Model's interruption reaction: \texttt{<predicted\_reaction>}

\vspace{0.4em}
\textbf{Interruption Type Definitions.}\\
False Trigger = mistaken interruption;

Cancellation = terminate; 

Correction = revise.

\vspace{0.4em}
\textbf{Evaluation Objective.}\\
Determine whether the model’s interruption reaction is consistent with its predicted interruption type.

\vspace{0.4em}
\textbf{Consistency Criteria.}
\begin{itemize}[leftmargin=1.4em,topsep=0.2em,itemsep=0.2em]
    \item \textbf{False Trigger}: The model should continue the original response path and should \emph{not} revise or terminate.
    \item \textbf{Correction}: The model should revise or reconstruct the original semantics, and should \emph{not} simply continue or directly terminate.
    \item \textbf{Cancellation}: The model should immediately terminate the task and should \emph{not} continue answering.
\end{itemize}

\textbf{Output Requirement.}
\begin{enumerate}[leftmargin=1.4em,topsep=0.2em,itemsep=0.2em]
    \item \textbf{Logical Analysis}: Briefly explain whether the reaction is logically consistent.
    \item \textbf{Final Score}:
    If logically consistent, output \texttt{Score: [1]}; 
    otherwise, output \texttt{Score: [0]}.
\end{enumerate}
\end{tcolorbox}

\noindent Textual Fluency Score asks the judge model to assess whether the model performs a natural and stable strategy transition after being interrupted, rather than exhibiting abrupt shifts, semantic discontinuities, or unnecessary repetition. All placeholders are filled with sample-specific information during inference.

\begin{tcolorbox}[
    colback=gray!4,
    colframe=gray!75,
    title={TFS Evaluation Prompt},
    fonttitle=\bfseries,
    boxrule=0.5pt,
    arc=1.5mm,
    left=2mm,right=2mm,top=1mm,bottom=1mm,
    toptitle=0.5mm,
    bottomtitle=0.5mm,
    before skip=5mm,
    after skip=1mm
]
\small

\textbf{Role.} 
You are a professional dialogue generation trajectory evaluation expert. You are assessing the \textbf{Textual Fluency Score} of a vision-language model when it performs a strategy shift after being interrupted by a user during a real-time video stream question-answering task.

\end{tcolorbox}

\begin{tcolorbox}[
    colback=gray!4,
    colframe=gray!75,
    title={TFS Evaluation Prompt (continued) },
    fonttitle=\bfseries,
    boxrule=0.5pt,
    arc=1.5mm,
    left=2mm,right=2mm,top=1mm,bottom=1mm,
    toptitle=0.5mm,
    bottomtitle=0.5mm,
    before skip=0mm,
    after skip=9mm
]
\small

\textbf{Metric Definition.}\\
The \textbf{Textual Fluency Score} metric evaluates whether the model achieves a natural and stable transition before and after the interruption, rather than exhibiting abrupt shifts, semantic discontinuities, or unnecessary repetition.

\vspace{0.4em}
\textbf{Interaction Background.}\\
Original question: \texttt{<question>} (timestamp: \texttt{<q\_ts>}s)\\
Response before interruption: \texttt{<truncated\_text>}\\
User interruption signal: \texttt{<interruption\_signal>} (timestamp: \texttt{<i\_ts>}s)

\vspace{0.4em}
\textbf{Model Behavior.}\\
Predicted interruption type: \texttt{<pred\_type>}\\
Model's interruption reaction: \texttt{<predicted\_reaction>}

\vspace{0.4em}
\textbf{Interruption Type Definitions.}\\
False Trigger = mistaken interruption; \\
Cancellation = terminate; \\
Correction = revise.

\vspace{0.4em}
\textbf{Evaluation Dimensions.}
\begin{enumerate}[leftmargin=1.4em,topsep=0.2em,itemsep=0.2em]
    \item \textbf{Boundary Continuity}: 
    Is the transition from the last sentence before interruption to the first sentence after interruption natural? 
    Is there any sudden topic shift, semantic jump, or syntactic break?

    \item \textbf{Controlled Strategy Shift}: 
    If False Trigger, does the model naturally continue along the original path? 
    If Correction, does the model smoothly shift toward revision rather than abruptly overturning prior content? 
    If Cancellation, does the model terminate cleanly rather than continuing with redundant output?

    \item \textbf{Unnecessary Discontinuity Detection}: 
    Is there repetition, disordered syntax, or logical collapse? 
    Is there obvious distributional collapse, such as suddenly generating completely irrelevant content?
\end{enumerate}

\vspace{0.4em}
\textbf{Scoring Criteria (1--10).}
Please assign a score from 1 to 10 based on transition smoothness, boundary continuity, strategy correctness, and the degree of unnecessary discontinuity. The detailed score-to-criterion mapping is summarized in Table~\ref{tab:sts_rubric}.

\vspace{0.4em}

\textbf{Output Requirement.}
\begin{itemize}[leftmargin=1.4em,topsep=0.2em,itemsep=0.2em]
    \item \textbf{Logical Analysis}: Specify which scoring rule(s) were triggered and how they apply.
    \item \textbf{Final Score}: \texttt{Score: [number]}
\end{itemize}
\end{tcolorbox}

\noindent Visual Evidence Score (VES) asks the judge model to verify whether the model’s post-interruption reaction is visually grounded in the corresponding video segment, with particular attention to visual authenticity, temporal synchronization, and hallucination detection. All placeholders are filled with sample-specific information during inference.

\begin{tcolorbox}[
    colback=gray!4,
    colframe=gray!75,
    title={VES Evaluation Prompt},
    fonttitle=\bfseries,
    boxrule=0.5pt,
    arc=1.5mm,
    left=2mm,right=2mm,top=1mm,bottom=1mm,
    toptitle=0.5mm,
    bottomtitle=0.5mm,
    before skip=8mm,
    after skip=8mm
]
\small

\textbf{Role.} 
You are a senior video content auditing expert. You are evaluating the \textbf{Visual Evidence Score} of a vision-language model during real-time video stream interaction.

\end{tcolorbox}

\begin{tcolorbox}[
    colback=gray!4,
    colframe=gray!75,
    title={VES Evaluation Prompt (continued) },
    fonttitle=\bfseries,
    boxrule=0.5pt,
    arc=1.5mm,
    left=2mm,right=2mm,top=1mm,bottom=1mm,
    toptitle=0.5mm,
    bottomtitle=0.5mm,
    before skip=0mm,
    after skip=7mm
]
\small

\textbf{Interaction Background.}\\
Original question: \texttt{<question>} (timestamp: \texttt{<q\_ts>}s)\\
Response before interruption: \texttt{<truncated\_text>}\\
User interruption signal: \texttt{<interruption\_signal>} (timestamp: \texttt{<i\_ts>}s)

\vspace{0.4em}
\textbf{Model Behavior.}\\
Predicted interruption type: \texttt{<pred\_type>}\\
Model's interruption reaction: \texttt{<predicted\_reaction>}

\vspace{0.4em}
\textbf{Core Evaluation Dimensions.}\\
You must not only examine the text, but also trace back to the visual frames between \texttt{<q\_ts>}s and \texttt{<i\_ts>}s to verify the following aspects:
\begin{enumerate}[leftmargin=1.4em,topsep=0.2em,itemsep=0.2em]
    \item \textbf{Visual Authenticity}: Do the objects, actions, colors, positions, and other details mentioned by the model actually appear in the video?
    \item \textbf{Timestamp Synchronization}: The interruption occurs at \texttt{<i\_ts>}s. Does the model’s reaction at that moment accurately reflect the visual state at that second, or within the preceding one or two seconds?
    \item \textbf{Hallucination Detection}: Did the model fabricate events or visual details that do not exist in the video in order to forcibly match the user’s interruption signal?
\end{enumerate}

\vspace{0.4em}
\textbf{Scoring Criteria (1--10).}\\
Please assign a score from 1 to 10 based on visual grounding accuracy, timestamp synchronization, and the degree of hallucination. 
The detailed score-to-criterion mapping is summarized in Table~\ref{tab:vra_rubric}.

\vspace{0.4em}
\textbf{Score Cap Conditions.}\\
Certain severe visual grounding errors impose upper bounds on the final score. 
The detailed cap conditions are summarized in Table~\ref{tab:vra_cap}.

\vspace{0.4em}
\textbf{Evaluation Task.}
\begin{itemize}[leftmargin=1.4em,topsep=0.2em,itemsep=0.2em]
    \item \textbf{Video Detail Verification}: What specific visual elements did the model mention? Which are genuinely present? Which are not? Which scoring rule(s) were triggered and how?
    \item \textbf{Final Score}: \texttt{Score: [number]}
\end{itemize}
\end{tcolorbox}

\noindent Intent Fulfillment Accuracy (IFA) asks the judge model to infer which implicit strategy the model adopted after being interrupted, and then determine whether this inferred strategy aligns with the target strategy implied by the ground-truth interruption type. All placeholders are filled with sample-specific information during inference.

\begin{tcolorbox}[
    colback=gray!4,
    colframe=gray!75,
    title={IFA Evaluation Prompt},
    fonttitle=\bfseries,
    boxrule=0.5pt,
    arc=1.5mm,
    left=2mm,right=2mm,top=1mm,bottom=1mm,
    toptitle=0.5mm,
    bottomtitle=0.5mm,
    before skip=5mm,
    after skip=0mm
]
\small

\textbf{Role.} 
You are an interruption strategy identification evaluation expert. Your task is to determine which implicit strategy the model adopted after being interrupted, and compare it with the target strategy.

\vspace{0.4em}
\textbf{Interaction Background.}\\
Original question: \texttt{<question>} (timestamp: \texttt{<q\_ts>}s)\\
Response before interruption: \texttt{<truncated\_text>}\\
User interruption signal: \texttt{<interruption\_signal>} (timestamp: \texttt{<i\_ts>}s)\\
GT interruption type: \texttt{<gt\_type>}

\end{tcolorbox}

\begin{tcolorbox}[
    colback=gray!4,
    colframe=gray!75,
    title={IFA Evaluation Prompt (continued) },
    fonttitle=\bfseries,
    boxrule=0.5pt,
    arc=1.5mm,
    left=2mm,right=2mm,top=1mm,bottom=1mm,
    toptitle=0.5mm,
    bottomtitle=0.5mm,
    before skip=0mm,
    after skip=9mm
]
\small

\textbf{Model Reaction After Interruption.}\\
\texttt{<predicted\_reaction>}

\vspace{0.4em}
\textbf{Definitions of the Three Strategy Behaviors.}
\begin{enumerate}[leftmargin=1.4em,topsep=0.2em,itemsep=0.2em]
    \item \textbf{Continuation (continue original path)}
    \begin{itemize}[leftmargin=1.4em,topsep=0.1em,itemsep=0.1em]
        \item Continues directly along the original response logic.
        \item Does not retract previous content.
    \end{itemize}

    \item \textbf{Revision (semantic correction)}
    \begin{itemize}[leftmargin=1.4em,topsep=0.1em,itemsep=0.1em]
        \item Explicitly retracts or negates the previous conclusion.
        \item Rewrites or corrects content under new constraints.
    \end{itemize}

    \item \textbf{Stop (termination)}
    \begin{itemize}[leftmargin=1.4em,topsep=0.1em,itemsep=0.1em]
        \item Explicitly stops the task.
        \item Outputs \texttt{[STOP]} or an equivalent termination expression.
    \end{itemize}
\end{enumerate}

\vspace{0.4em}
\textbf{Evaluation Procedure.}
\begin{enumerate}[leftmargin=1.4em,topsep=0.2em,itemsep=0.2em]
    \item Determine which category the model’s actual behavior belongs to.
    \item Compare it with the target strategy corresponding to the GT interruption type:
    \begin{itemize}[leftmargin=1.4em,topsep=0.1em,itemsep=0.1em]
        \item \textbf{False Trigger} $\Rightarrow$ \textbf{Continuation} (continue the original response)
        \item \textbf{Correction} $\Rightarrow$ \textbf{Revision} (rewrite under new constraints)
        \item \textbf{Cancellation} $\Rightarrow$ \textbf{Stop} (terminate the task, output \texttt{[STOP]})
    \end{itemize}
    \item Assign the final score.

\end{enumerate}

\vspace{0.4em}
\textbf{Special Rules.}
\begin{itemize}[leftmargin=1.4em,topsep=0.2em,itemsep=0.2em]
    \item If the model first continues the old conclusion and then performs correction, count it as incorrect.
    \item If the model does not explicitly stop and continues generating content, it does \emph{not} qualify as \textbf{Stop}.
    \item If the behavior is ambiguous or mixes two strategies, count it as incorrect.
\end{itemize}

\vspace{0.4em}
\textbf{Output Requirement.}
\begin{itemize}[leftmargin=1.4em,topsep=0.2em,itemsep=0.2em]
    \item \textbf{Logical Analysis}: Clearly explain the inferred strategy and the reasoning, and explicitly state whether it aligns with the target strategy.
    \item \textbf{Final Score}: Output \texttt{Score: [1]} if consistent; otherwise output \texttt{Score: [0]}.
\end{itemize}
\end{tcolorbox}

\noindent Correction Following Score (CFS) asks the judge model to assess whether, under a \textbf{Correction} scenario, the model truly performs semantic retraction and structural reconstruction, rather than merely applying superficial paraphrasing. 
All placeholders are filled with sample-specific information during inference.

\begin{tcolorbox}[
    colback=gray!4,
    colframe=gray!75,
    title={CFS Evaluation Prompt},
    fonttitle=\bfseries,
    boxrule=0.5pt,
    arc=1.5mm,
    left=2mm,right=2mm,top=1mm,bottom=1mm,
    toptitle=0.5mm,
    bottomtitle=0.5mm,
    before skip=5mm,
    after skip=9mm
]
\small

\textbf{Role.} 
You are a professional semantic revision quality evaluation expert. You only evaluate \textbf{Correction} scenarios, assessing whether the model truly performs semantic retraction and structural reconstruction, rather than superficial paraphrasing.

\vspace{0.4em}
\textbf{Scoring Objective.}\\
If the GT interruption type is \emph{not} \textbf{Correction}, directly output \texttt{Score: [0]}. 
If the GT interruption type is \textbf{Correction}, determine whether the model genuinely retracts the old conflicting semantics and reconstructs the response under the new constraints.

\vspace{0.4em}
\textbf{Interaction Background.}\\
Original question: \texttt{<question>} (timestamp: \texttt{<q\_ts>}s)\\
Response before interruption: \texttt{<truncated\_text>}\\
User interruption signal: \texttt{<interruption\_signal>} (timestamp: \texttt{<i\_ts>}s)

\vspace{0.4em}
\textbf{Model Behavior.}\\
Predicted interruption type: \texttt{<pred\_type>}\\
Model's interruption reaction: \texttt{<predicted\_reaction>}

\vspace{0.4em}
\textbf{Evaluation Procedure (must follow in order).}
\begin{enumerate}[leftmargin=1.4em,topsep=0.2em,itemsep=0.2em]
    \item \textbf{Identify the conflict points in the old semantics.}
    \begin{itemize}[leftmargin=1.4em,topsep=0.1em,itemsep=0.1em]
        \item Which statements contradict the user’s correction signal?
        \item Which conclusions must be retracted?
    \end{itemize}

    \item \textbf{Check whether the model:}
    \begin{itemize}[leftmargin=1.4em,topsep=0.1em,itemsep=0.1em]
        \item Explicitly retracts or negates the conflicting content
        \item Reconstructs the semantics under the new constraints
        \item Avoids continuing to use the old conclusions
    \end{itemize}
\end{enumerate}

\vspace{0.4em}
\textbf{Scoring Criteria (1--10).}\\
Please assign a score from 1 to 10 according to the detailed semantic revision rubric summarized in Table~\ref{tab:sra_rubric}.

\vspace{0.4em}
\textbf{Score Cap Rules.}\\
Certain severe semantic revision failures impose upper bounds on the final score. 
The detailed cap rules are summarized in Table~\ref{tab:sra_cap}.

\vspace{0.4em}
\textbf{Output Requirement.}
\begin{itemize}[leftmargin=1.4em,topsep=0.2em,itemsep=0.2em]
    \item \textbf{Logical Analysis}: Identify the old semantic conflict points, evaluate whether the new constraints are satisfied, determine whether old conflicts remain, and specify which scoring rule(s) were triggered and how.
    \item \textbf{Final Score}: \texttt{Score: [number]}
\end{itemize}
\end{tcolorbox}

\noindent Contextual Consistency Score (CCS) asks the judge model to assess whether the model preserves valid and unaffected context after an interruption, while still performing any necessary modification or termination. 
All placeholders are filled with sample-specific information during inference.

\begin{tcolorbox}[
    colback=gray!4,
    colframe=gray!75,
    title={CCS Evaluation Prompt},
    fonttitle=\bfseries,
    boxrule=0.5pt,
    arc=1.5mm,
    left=2mm,right=2mm,top=1mm,bottom=1mm,
    toptitle=0.5mm,
    bottomtitle=0.5mm,
    before skip=5mm,
    after skip=9mm
]
\small

\textbf{Role.} 
You are a context preservation capability evaluation expert. Your task is to assess whether the model preserves unaffected valid context after an interruption, while also performing necessary modification or termination.

\vspace{0.4em}
\textbf{Interaction Background.}\\
Original question: \texttt{<question>} (timestamp: \texttt{<q\_ts>}s)\\
Response before interruption: \texttt{<truncated\_text>}\\
User interruption signal: \texttt{<interruption\_signal>} (timestamp: \texttt{<i\_ts>}s)

\vspace{0.4em}
\textbf{Model Behavior.}\\
Predicted interruption type: \texttt{<pred\_type>}\\
Model's interruption reaction: \texttt{<predicted\_reaction>}

\vspace{0.4em}
\textbf{Scoring Focus.}
\begin{enumerate}[leftmargin=1.4em,topsep=0.2em,itemsep=0.2em]
    \item \textbf{Preservation}: Are key information or context units that should remain unaffected by the interruption preserved, such as characters, events or goals?
    \item \textbf{Precise Modification}: Were the necessary parts correctly modified, especially in \textbf{Correction}, while unaffected parts were not unnecessarily altered?
\end{enumerate}

\vspace{0.4em}
\textbf{Evaluation Procedure (must follow in order).}
\begin{enumerate}[leftmargin=1.4em,topsep=0.2em,itemsep=0.2em]
    \item \textbf{Identify the semantic units that should be preserved.}
    \begin{itemize}[leftmargin=1.4em,topsep=0.1em,itemsep=0.1em]
        \item Established characters or objects
        \item Previously described facts
        \item Context information unrelated to the correction
    \end{itemize}

    \item \textbf{Examine the post-interruption reaction.}
    \begin{itemize}[leftmargin=1.4em,topsep=0.1em,itemsep=0.1em]
        \item Which semantic units were preserved?
        \item Which were unnecessarily removed?
        \item Is there evidence of complete overturning and rewriting?
    \end{itemize}

    \item \textbf{Determine whether any of the following occur.}
    \begin{itemize}[leftmargin=1.4em,topsep=0.1em,itemsep=0.1em]
        \item Unnecessary large-scale rewriting
        \item Complete replacement of the original context
    \end{itemize}
\end{enumerate}

\vspace{0.4em}
\textbf{Scoring Criteria (1--10).}\\
Please assign a score from 1 to 10 according to the detailed context preservation rubric summarized in Table~\ref{tab:cpi_rubric}.

\vspace{0.4em}
\textbf{Score Cap Rules.}\\
Certain severe context preservation failures impose upper bounds on the final score. 
The detailed cap rules are summarized in Table~\ref{tab:cpi_cap}.

\vspace{0.4em}
\textbf{Output Requirement.}
\begin{itemize}[leftmargin=1.4em,topsep=0.2em,itemsep=0.2em]
    \item \textbf{Logical Analysis}: List the semantic units that should be preserved, specify which were preserved or removed, and indicate which scoring rule(s) were triggered and how.
    \item \textbf{Final Score}: \texttt{Score: [number]}
\end{itemize}
\end{tcolorbox}

\clearpage

\begin{table*}[t]
\centering
\small
\caption{Detailed scoring rubric for Textual Fluency Score (TFS).}
\label{tab:sts_rubric}
\setlength{\tabcolsep}{5pt}
\renewcommand{\arraystretch}{1.3}
\begin{tabular}{>{\centering\arraybackslash}m{0.11\linewidth} m{0.82\linewidth}}
\toprule
\rowcolor[HTML]{F3F3F3}
\textbf{Score} & \multicolumn{1}{c}{\textbf{Criterion}} \\
\midrule
10 & Immediately enters the correct strategy; no repetition, filler transition, or semantic jump; reads like natural human real-time dialogue. \\
\rowcolor[HTML]{FAFAFA}
9 & Correct strategy executed immediately; may contain a very short transition word; no repetition or semantic discontinuity. \\
8 & Strategy is correct; may contain one explicit transition sentence; no semantic confusion; no repetition exceeding one sentence. \\
\rowcolor[HTML]{FAFAFA}
7 & Strategy is correct, with only minor repetition or slightly rigid structure. \\
6 & Strategy is correct, but repeats original content for more than one sentence or briefly becomes illogical before recovering. \\
\rowcolor[HTML]{FAFAFA}
5 & Eventually shifts to the correct strategy, but contains noticeable discontinuity or delayed revision. \\
4 & Strategy execution is unstable, with large repetition. \\
\rowcolor[HTML]{FAFAFA}
3 & Weak connection to prior content; semantic jump or topic drift occurs. \\
2 & Post-interruption content is almost disconnected from prior context; logical confusion is obvious. \\
\rowcolor[HTML]{FAFAFA}
1 & Completely ignores the interruption or generates content severely detached from context. \\
\bottomrule
\end{tabular}
\end{table*}

\begin{table*}[t]
\centering
\small
\caption{Detailed scoring rubric for Visual Evidence Score (VES).}
\label{tab:vra_rubric}
\setlength{\tabcolsep}{5pt}
\renewcommand{\arraystretch}{1.3}
\begin{tabular}{>{\centering\arraybackslash}m{0.11\linewidth} m{0.82\linewidth}}
\toprule
\rowcolor[HTML]{F3F3F3}
\textbf{Score} & \multicolumn{1}{c}{\textbf{Criterion}} \\
\midrule
10 & All mentioned objects, actions, and attributes (e.g., color, position, state) are genuinely present; no hallucination; the description accurately reflects the visual state at the interruption moment; no temporal misalignment. \\
\rowcolor[HTML]{FAFAFA}
9 & All visual elements genuinely exist; only extremely minor detail imprecision is present; timing is essentially synchronized. \\
8 & Major visual elements are correct; one minor detail may be vague or slightly inaccurate; no obvious hallucination. \\
\rowcolor[HTML]{FAFAFA}
7 & Core objects and actions are correct, with only one minor visual error or slight temporal lag. \\
6 & Most visual content is correct, but there is one obvious but non-core visual error, or a noticeable temporal synchronization deviation. \\
\rowcolor[HTML]{FAFAFA}
5 & Core visual elements are correct, but multiple minor visual errors are present, or there is partial over-generalization. \\
4 & One core visual error is present, or there is severe temporal misalignment, such as referring to frames several seconds earlier. \\
\rowcolor[HTML]{FAFAFA}
3 & Multiple core visual errors are present. \\
2 & Major visual information is severely inconsistent with the actual video. \\
\rowcolor[HTML]{FAFAFA}
1 & The content is unrelated to the video, describes a completely fabricated scene, or exhibits severe visual hallucination. \\
\bottomrule
\end{tabular}
\end{table*}

\begin{table*}[t]
\centering
\small
\caption{Detailed scoring rubric for Correction Following Score (CFS).}
\label{tab:sra_rubric}
\setlength{\tabcolsep}{5pt}
\renewcommand{\arraystretch}{1.3}
\begin{tabular}{>{\centering\arraybackslash}m{0.11\linewidth} m{0.82\linewidth}}
\toprule
\rowcolor[HTML]{F3F3F3}
\textbf{Score} & \multicolumn{1}{c}{\textbf{Criterion}} \\
\midrule
10 & Explicitly states that the old conclusion was incorrect; fully retracts the conflicting semantics; completely reconstructs the response under the new constraints; no residual old semantics remain. \\
\rowcolor[HTML]{FAFAFA}
9 & Fully follows the new constraints; does not explicitly apologize or deny the old conclusion, but contains no actual residual semantic error. \\
8 & Major semantic conflicts are corrected; only extremely minor and irrelevant old phrasing remains. \\
\rowcolor[HTML]{FAFAFA}
7 & The core conflict is corrected, but some secondary old semantics remain uncleared. \\
6 & The model attempts correction, but still retains noticeable fragments of the old conclusion. \\
\rowcolor[HTML]{FAFAFA}
5 & The model performs partial rewriting, but the core constraints are not fully satisfied. \\
4 & The model only superficially modifies wording, while substantively continuing the old semantics. \\
\rowcolor[HTML]{FAFAFA}
3 & The response clearly mixes old and new conclusions, and logical conflict remains. \\
2 & The model largely fails to retract the old semantics and remains clearly inconsistent with the new constraints. \\
\rowcolor[HTML]{FAFAFA}
1 & The model completely ignores the correction signal or generates content unrelated to the task. \\
\bottomrule
\end{tabular}
\end{table*}

\begin{table*}[t]
\centering
\small
\caption{Detailed scoring rubric for Contextual Consistency Score (CCS).}
\label{tab:cpi_rubric}
\setlength{\tabcolsep}{5pt}
\renewcommand{\arraystretch}{1.5}
\begin{tabular}{>{\centering\arraybackslash}m{0.11\linewidth} m{0.82\linewidth}}
\toprule
\rowcolor[HTML]{F3F3F3}
\textbf{Score} & \multicolumn{1}{c}{\textbf{Criterion}} \\
\midrule
10 & All semantic units that should be preserved are fully retained; only minimal necessary modifications are made; no irrelevant expansion or structural disruption occurs. \\
\rowcolor[HTML]{FAFAFA}
9 & Almost all preservable context is retained; only a very small amount of non-core information is lost. \\
8 & All core semantics are preserved, but some secondary details are missing. \\
\rowcolor[HTML]{FAFAFA}
7 & The main semantics are preserved, but some core information is slightly weakened. \\
6 & Core semantics are only partially preserved, and some unrelated content is clearly deleted or rewritten. \\
\rowcolor[HTML]{FAFAFA}
5 & Only part of the core semantics is preserved, and the structure is noticeably reorganized. \\
4 & A large amount of originally valid semantics is deleted, with clear overturn-and-rewrite behavior. \\
\rowcolor[HTML]{FAFAFA}
3 & Only very little original semantics is preserved; the response essentially regenerates a new version of the content. \\
2 & The original context is almost erased, and the semantic structure is largely rebuilt. \\
\rowcolor[HTML]{FAFAFA}
1 & The model completely fails to preserve the original valid context and becomes detached from the original content. \\
\bottomrule
\end{tabular}
\end{table*}

\begin{table*}[t]
\centering
\caption{Score cap conditions for Visual Evidence Score (VES).}
\label{tab:vra_cap}
\setlength{\tabcolsep}{5pt}
\renewcommand{\arraystretch}{1.5}
\begin{tabular}{m{0.56\linewidth} >{\centering\arraybackslash}m{0.4\linewidth}}
\toprule
\rowcolor[HTML]{F3F3F3}
\multicolumn{1}{c}{\textbf{Condition}} & \textbf{Maximum Score} \\
\midrule
\multicolumn{1}{c}{One core visual error} & $\leq 4$ \\
\rowcolor[HTML]{FAFAFA}
\multicolumn{1}{c}{Two or more core visual errors} & $\leq 3$ \\
\multicolumn{1}{c}{Obvious fabricated scene} & $\leq 2$ \\
\rowcolor[HTML]{FAFAFA}
\multicolumn{1}{c}{Content completely detached from the video} & $1$ \\
\bottomrule
\end{tabular}
\end{table*}

\begin{table*}[t]
\centering
\caption{Score cap rules for Correction Following Score (CFS).}
\label{tab:sra_cap}
\setlength{\tabcolsep}{5pt}
\renewcommand{\arraystretch}{1.5}
\begin{tabular}{p{0.6\linewidth} >{\centering\arraybackslash}p{0.45\linewidth}}
\toprule
\rowcolor[HTML]{F3F3F3}
\multicolumn{1}{c}{\textbf{Condition}} & \textbf{Maximum Score} \\
\midrule
\multicolumn{1}{c}{Clearly continues the core old conclusion} & $\leq 4$ \\
\rowcolor[HTML]{FAFAFA}
\multicolumn{1}{c}{Old and new conclusions coexist} & $\leq 3$ \\
\multicolumn{1}{c}{Completely ignores the correction signal} & $1$ \\
\bottomrule
\end{tabular}
\end{table*}

\begin{table*}[t]
\centering
\caption{Score cap rules for Contextual Consistency Score (CCS).}
\label{tab:cpi_cap}
\setlength{\tabcolsep}{5pt}
\renewcommand{\arraystretch}{1.5}
\begin{tabular}{p{0.6\linewidth} >{\centering\arraybackslash}p{0.4\linewidth}}
\toprule
\rowcolor[HTML]{F3F3F3}
\multicolumn{1}{c}{\textbf{Condition}} & \textbf{Maximum Score} \\
\midrule
\multicolumn{1}{c}{Unjustified deletion of core semantic units} & $\leq 4$ \\
\rowcolor[HTML]{FAFAFA}
\multicolumn{1}{c}{Large-scale rewriting of original context} & $\leq 3$ \\
\multicolumn{1}{c}{Complete topic shift} & $1$ \\
\bottomrule
\end{tabular}
\end{table*}

\clearpage

\section{Data Format}

Following the task formulation of Online Video Question Answering under Interruption, each benchmark instance is organized around a temporally grounded interrupted interaction. OVIBench provides two complementary data formats: an open-ended format for free-form response evaluation and a multiple-choice format for controlled, low-noise assessment.

\subsection{Open-Ended Format}

The open-ended format is designed to approximate realistic interrupted interactions, where the model must generate a free-form response after receiving an interruption during answer generation. Each sample contains the original question, the timestamp when the question is issued, the interruption signal, the timestamp when the interruption occurs, and the source video.

For implementation, each sample is stored together with a unique identifier and the interruption type. The question and interruption signal are grouped under an interaction setup field, which explicitly specifies their timestamps. An example is shown below.

\vspace{0.3em}

\begin{lstlisting}[style=jsonstyle]
{
  "case_id": "ActivityNet_120.mp4_s0_Cancellation",
  "video_file": "DATA/my_large_dataset/ActivityNet_120.mp4",
  "interruption_type": "Cancellation",
  "interaction_setup": {
    "question": "Describe the visual elements and possible function ...",
    "question_timestamp": 0.97,
    "interruption_signal": "Stop talking",
    "interruption_timestamp": 6.13
  }
}
\end{lstlisting}

\vspace{0.3em}

In this example, the model starts answering the original question at timestamp 0.97 and then receives a \texttt{Cancellation} interruption at timestamp 6.13. During evaluation, the model is expected to respond according to the interruption signal under this temporal setup.

\subsection{Multiple-Choice Format}

To reduce evaluation noise and support more reproducible comparisons, OVIBench also provides a multiple-choice version of each interrupted interaction. Instead of requiring the model to freely generate a response, this format presents several candidate answers and asks the model to choose the one that best matches the interruption context. In addition, we avoid constructing overly simple distractors in the multiple-choice setting. Instead of using random irrelevant answers, the incorrect option is designed to be close to plausible responses in surface form but wrong in key semantics.

As shown in the example below, each sample records the interaction state at the moment of interruption, including the dialogue history, the current question, the partial answer generated before interruption, the interruption signal and timestamp, and the corresponding visual stream range. In addition, each sample contains four answer options and a ground-truth label. To avoid positional bias, the option order is randomly shuffled in each sample. An example is given below.

\vspace{0.6em}
\begin{lstlisting}[style=jsonstyle]
{
  "case_id": "ActivityNet_25.mp4_q0_Cancellation",
  "input_state": {
    "history_qa": [],
    "current_question": "Are there any tents or structures visible ...",
    "question_timestamp": 5.81,
    "partial_answer_generated": "Yes, in the background ...",
    "interruption_signal": "That's enough",
    "interruption_timestamp": 6.77,
    "visual_stream_range": [5.81, 6.77]
  },
  "options": {
    "A": "...",
    "B": "Understood, I have stopped ...",
    "C": "...",
    "D": "[STOP]"
  },
  "ground_truth": {
    "type": "Cancellation",
    "label": "D",
    "content": "[STOP]",
    "behavior": "stop"
  }
}
\end{lstlisting}
\vspace{0.6em}

In this example, the interruption signal ``That's enough'' indicates a \texttt{Cancellation} case. Since the model is expected to stop responding after this interruption, the correct choice is \texttt{[STOP]}. In this way, the multiple-choice format complements the open-ended setting by providing a more stable benchmark for interruption understanding and response selection.

\section{Full Experimental Results}

This section presents the full open-ended results of OVIBench for the three interruption types: \texttt{Cancellation}, \texttt{Correction}, and \texttt{False Trigger}. These tables provide a more detailed breakdown of model behavior across all metrics and further support the main findings that interruption handling is strongly type-dependent: \texttt{Cancellation} is relatively easy to recognize, \texttt{Correction} remains challenging in precise response revision, and \texttt{False Trigger} shows the largest performance variation across models.

\begin{table*}[t!]
\centering
\caption{Experimental results on the open-ended subset of OVIBench for Cancellation. The best results in each column are highlighted in bold.}
\setlength{\tabcolsep}{10pt} 
\renewcommand{\arraystretch}{1.7} 

\begin{tabular}{l*{7}{c}} 
\toprule
\rowcolor[HTML]{F9F9F9} \textbf{Model/Metric} & \textbf{IAC} & \textbf{TFS} & \textbf{VES} & \textbf{IFA} & \textbf{CFS} & \textbf{CCS} & \textbf{CA} \\
\midrule
doubao & \textbf{95.54\%} & 9.35 & 7.18 & 64.53\% & 1.69 & 5.80 & \textbf{100\%} \\
\rowcolor[HTML]{F2F2F2} 
gemini & 79.94\% & \textbf{9.47} & \textbf{7.49} & 37.90\% & 2.20 & 5.13 & \textbf{100\%}  \\
Qwen2.5-VL-7B & 51.45\% & 7.55 & 4.83 & 27.76\% & 1.34 & 4.56 & \textbf{100\%}  \\
\rowcolor[HTML]{F2F2F2} 
Qwen3-VL-8B & 90.40\% & 9.13 & 6.80 & 67.79\% & 1.85 & 6.55 & \textbf{100\%}  \\
VideoChat-R1-7B & 58.24\% & 7.81 & 5.03 & 36.79\% & 1.33 & 4.26 & \textbf{100\%}  \\
\rowcolor[HTML]{F2F2F2} 
VideoLLaMA3-7B & 74.17\% & 7.33 &4.99 & 40.38\% & 1.51 & 5.26 & \textbf{100\%}  \\
qwen3-30b-a3b & 88.37\% & 9.34 & 6.62 & \textbf{69.92\%} & \textbf{2.23} & \textbf{6.77} & \textbf{100\%} \\
\bottomrule
\end{tabular}

\end{table*}

\begin{table*}[t!]
\centering
\caption{Experimental results on the open-ended subset of OVIBench for Correction. The best results in each column are highlighted in bold.}
\setlength{\tabcolsep}{10pt} 
\renewcommand{\arraystretch}{1.6} 

\begin{tabular}{l*{7}{c}} 
\toprule
\rowcolor[HTML]{F9F9F9} \textbf{Model/Metric} & \textbf{IAC} & \textbf{TFS} & \textbf{VES} & \textbf{IFA} & \textbf{CFS} & \textbf{CCS} & \textbf{CA} \\
\midrule
doubao & 95.33\% & \textbf{9.81} & 6.57 & 78.48\% & \textbf{8.73} & 7.22 & 88\% \\
\rowcolor[HTML]{F2F2F2} 
gemini & 69.62\% & 8.02 & 6.42 & 67.94\% & 7.14 & \textbf{7.33} & 87\%  \\
Qwen2.5-VL-7B & 92.92\% & 9.02 & 6.63 & 77.14\% & 6.49 & 5.79 & \textbf{92\%}  \\
\rowcolor[HTML]{F2F2F2} 
Qwen3-VL-8B & \textbf{98.53\%} & 9.77 & \textbf{7.28} & 89.96\% & 8.25 & 6.30 & 87\%  \\
VideoChat-R1-7B & 94.18\% & 9.16 & 6.65 & 74.59\% & 6.76 & 5.73 & \textbf{92\%}  \\
\rowcolor[HTML]{F2F2F2} 
VideoLLaMA3-7B & 97.63\% & 9.06 & 6.14 & 66.25\% & 7.35 & 6.15 & 90\%  \\
qwen3-30b-a3b & 98.47\% & 9.68 & 7.03 & \textbf{91.04\%} & 8.68 & 6.09 & 89\% \\
\bottomrule
\end{tabular}

\end{table*}

\begin{table*}[t!]
\centering
\caption{Experimental results on the open-ended subset of OVIBench for False Trigger. The best results in each column are highlighted in bold.}
\setlength{\tabcolsep}{10pt} 
\renewcommand{\arraystretch}{1.6} 

\begin{tabular}{l*{7}{c}} 
\toprule
\rowcolor[HTML]{F9F9F9} \textbf{Model/Metric} & \textbf{IAC} & \textbf{TFS} & \textbf{VES} & \textbf{IFA} & \textbf{CFS} & \textbf{CCS} & \textbf{CA} \\
\midrule
doubao & \textbf{96.74\%} & \textbf{9.49} & 4.55 & \textbf{96.83\%} & \textbf{1.92} & \textbf{8.87} & 98\% \\
\rowcolor[HTML]{F2F2F2} 
gemini & 86.51\% & 8.10 & 4.73 & 85.64\% & 1.30 & 8.15 & \textbf{99\%}  \\
Qwen2.5-VL-7B & 19.62\% & 4.60 & 3.27 & 19.16\% & 0.69 & 4.12 & 67\%  \\
\rowcolor[HTML]{F2F2F2} 
Qwen3-VL-8B & 87.19\% & 8.62 & \textbf{5.35} & 77.06\% & 1.41 & 7.88 & 85\%  \\
VideoChat-R1-7B & 19.25\% & 4.61 & 3.09 & 18.60\% & 0.69 & 4.21 & 75\%  \\
\rowcolor[HTML]{F2F2F2} 
VideoLLaMA3-7B & 34.52\% & 6.35 & 5.28 & 48.27\% & 0.91 & 5.70 & 84\%  \\
qwen3-30b-a3b & 67.85\% & 7.50 & 5.33 & 69.28\% & 0.94 & 6.90 & 90\% \\
\bottomrule
\end{tabular}
\end{table*}

\begin{figure*}[t]
\centering
\includegraphics[width=\linewidth]{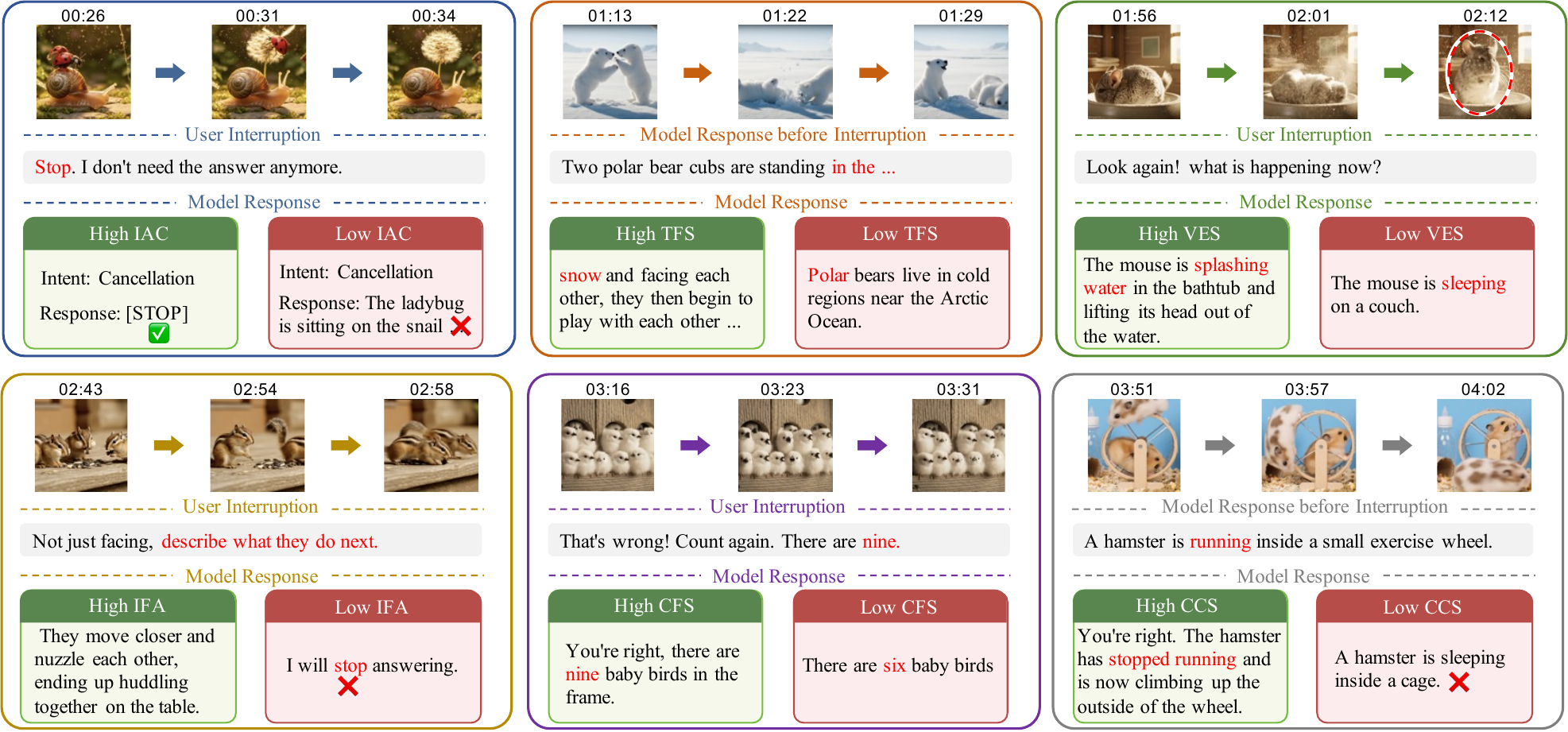}
\caption{Illustrative examples of the six evaluation metrics in OVIBench. Each case shows a short video stream with a user interruption and the model's responses. Green boxes denote responses that achieve high scores, while red boxes indicate failure cases.
}
\label{fig:index}
\end{figure*}

\section{Visualization of Evaluation Metrics}
Figure~\ref{fig:index} provides intuitive examples of the six evaluation metrics used in OVIBench. Each case contains a short video stream, a user interruption, and representative model responses. Green boxes indicate high-quality responses, while red boxes show typical failure cases.

The figure illustrates how different metrics focus on different aspects of interruption handling, including whether the model follows the interruption intent, maintains textual fluency, grounds its response in visual evidence, satisfies the user's request, follows correction signals, and preserves valid context. This visualization complements the metric definitions and helps clarify the scoring criteria used in human and model-based evaluation.

\section{Human Evaluation Subset}

To further verify whether the observed model performance and the gains from interruption-aware fine-tuning are stable beyond automatically generated data, we construct a human-written evaluation subset, denoted as OVI-Human. This subset is not used for OVI-Train training, prompt tuning, or model selection. Instead, it is only used as an additional evaluation set to examine whether the main conclusions still hold under manually written questions and interruption expressions.

OVI-Human contains 180 samples, with 60 samples for each interruption type: Cancellation, False Trigger, and Correction. The questions cover diverse user intents, including video description, object counting, action change, visual detail localization, and simple causal reasoning. The interruption signals are also written in more natural forms, such as implicit stop requests, colloquial corrections, and irrelevant short insertions. The answer options are manually written or substantially rewritten to reduce the style bias of automatic generation. We evaluate models on OVI-Human using the same multiple-choice metrics as the main experiment, namely Multiple-Choice Accuracy (MA) and Interruption-Type Classification Accuracy (CA).

\begin{table*}[t]
\centering
\caption{Multiple-choice evaluation results on the human-written OVI-Human subset. MA denotes answer selection accuracy, and CA denotes interruption-type classification accuracy. $\Delta$ indicates the performance improvement of our model over the baseline. The best results in each column are highlighted in bold.}
\label{tab:human_holdout_results}

\resizebox{0.95\textwidth}{!}{
\renewcommand{\arraystretch}{1.3}
\begin{tabular}{lcccccccc}
\toprule
\textbf{Model} & \multicolumn{2}{c}{\textbf{All}} & \multicolumn{2}{c}{\textbf{False Trigger}} & \multicolumn{2}{c}{\textbf{Cancellation}} & \multicolumn{2}{c}{\textbf{Correction}} \\
\cline{2-9}
& \textbf{MA} & \textbf{CA} & \textbf{MA} & \textbf{CA} & \textbf{MA} & \textbf{CA} & \textbf{MA} & \textbf{CA} \\
\midrule
\rowcolor[HTML]{F0F0F0} 
\multicolumn{9}{c}{\textit{Open-Source Models}} \\
\midrule
Qwen2.5-VL-7B~\cite{yao2024fine} 
& 62.78 & 79.44 & 48.33 & 91.67 & 42.78 & 50.00 & 85.56 & 94.67 \\
Qwen2.5-VL-72B~\cite{yao2024fine} 
& 69.44 & 90.56 & 86.67 & 96.67 & 38.33 & \textbf{91.67} & 83.33 & 83.33 \\
VideoChat-R1-7B~\cite{li2025videochat} 
& 68.33 & 81.67 & 52.78 & 91.67 & 49.44 & 58.33 & 88.89 & 95.00 \\
Qwen3-VL-30B-A3B~\cite{bai2025qwen3} 
& 70.00 & 86.11 & 28.33 & 73.33 & 68.89 & 81.67 & 96.67 & 98.33 \\
\rowcolor[HTML]{F0F0F0} 
\midrule
\multicolumn{9}{c}{\textit{Closed-Source Models}} \\
\midrule
Doubao-Seed-1.6~\cite{huang2025memorb}  
& 69.44 & 92.22 & 38.89 & 95.00 & 51.67 & 83.33 & 97.78 & 96.33 \\
Gemini-2.5-Flash~\cite{comanici2025gemini} 
& 55.00 & 66.11 & 60.00 & 68.33 & 21.67 & 53.33 & 83.33 & 76.67 \\
\midrule
\textbf{Ours (7B)} 
& \textbf{80.56} & \textbf{94.44} 
& \textbf{90.00} & \textbf{98.33} 
& \textbf{76.67} & 88.33
& \textbf{97.78} & \textbf{96.67} \\
\textbf{$\Delta$ (vs Qwen2.5-VL-7B)} 
& +17.78 & +15.00 
& +41.67 & +6.66 
& +33.89 & +38.33 
& +12.22 & +2.00 \\
\bottomrule
\end{tabular}
}
\end{table*}

As shown in Table~\ref{tab:human_holdout_results}, the results on OVI-Human show a similar overall trend to the main multiple-choice evaluation. Our fine-tuned model still achieves the best overall performance, improving All MA by 17.78 points and All CA by 15.00 points over Qwen2.5-VL-7B. The improvements are especially clear on False Trigger and Cancellation, indicating that OVI-Train helps the model better distinguish ineffective interruptions and stopping requests. These results suggest that the gains from interruption-aware fine-tuning are not limited to automatically generated samples. Since OVI-Human uses manually written questions and more natural interruption expressions, the consistent improvement provides additional evidence that the model learns useful interruption-handling ability rather than only fitting the style of the generated benchmark.

\section{Human Evaluation Instructions}
To ensure the quality of the OVIBench dataset, we conducted a human evaluation after the automatic data construction process. OVIBench focuses on online video question answering under interruption, where a model is expected to react properly to user interruptions during answer generation. The dataset contains three interruption types: \textit{Cancellation}, \textit{False Trigger}, and \textit{Correction}, corresponding to stopping the response, ignoring an ineffective interruption, and revising the response according to new user-provided information.

We invited 10 human participants to evaluate the generated samples. Each participant was asked to inspect the video content, the original question, the interruption signal, the interruption type label, the partial response before interruption, and the post-interruption response. For multiple-choice samples, participants also checked whether the candidate options and the ground-truth answer were reasonable. The final human score of each sample was computed by averaging the scores from the 10 participants, which reduces the influence of individual subjective bias.

The human evaluation mainly considered the following aspects:
\begin{enumerate}
    \item \textbf{Correctness of interruption type.} Participants checked whether the interruption signal was consistent with the annotated interruption type. For example, a stop request should be labeled as \textit{Cancellation}, a corrective signal should be labeled as \textit{Correction}, and an irrelevant or accidental signal should be labeled as \textit{False Trigger}.

    \item \textbf{Appropriateness of response behavior.} Participants examined whether the post-interruption response matched the expected behavior. A \textit{Cancellation} case should stop the response, a \textit{False Trigger} case should continue the original answer, and a \textit{Correction} case should revise the previous response according to the user correction.

    \item \textbf{Consistency with visual evidence.} Participants checked whether the objects, actions, scenes, and temporal information mentioned in the response were supported by the video content, avoiding obvious hallucinations or contradictions.

    \item \textbf{Contextual coherence.} Participants evaluated whether the response before and after the interruption was coherent. For correction cases, the model should revise the incorrect part while preserving valid context that is not affected by the correction.
    
    \item \textbf{Quality of multiple-choice options.} For multiple-choice samples, participants checked whether the correct option was unique and consistent with the interruption context, and whether the distractors were distinguishable without being equally correct.
\end{enumerate}

For scoring, binary criteria, such as whether the response behavior is correct, were marked as 0 or 1. Other quality dimensions, including textual fluency, visual consistency, correction quality, and contextual coherence, were scored on a 1--10 scale. For each sample, we collected the scores from all 10 participants and used their average as the final human evaluation score. In this way, the final score reflects the overall judgment of multiple human evaluators rather than the opinion of a single participant.
Samples with low scores or large disagreement among participants were manually rechecked. According to the identified problem, these samples were removed, revised, or relabeled. Through this human evaluation process, we filtered out samples with incorrect interruption labels, inconsistent visual evidence, unreasonable response logic, or ambiguous answer options, thereby improving the overall quality and reliability of the dataset.

\section{Offline Simulation Pseudocode}

To enable large-scale and reproducible evaluation, we simulate the online interruption process offline under a unified temporal protocol. As described in the main paper, the simulation first determines the observable video segment at the question timestamp, then generates a preliminary answer, truncates it according to the interruption time and a fixed generation rate, and finally feeds the interruption signal together with the updated visual context back to the model for post-interruption response generation. In this way, the protocol approximates interruption during answer generation while remaining suitable for batch evaluation and controlled comparison across models.

\vfill\break

\captionsetup[algorithm]{labelformat=empty}
\begin{algorithm}[H]
\caption{\textbf{Offline Simulation of Online Video Interruption}}
\label{alg:offline_simulation}
\begin{algorithmic}
\Require Target model $\mathcal{M}$, benchmark set $\mathcal{B}$, character generation rate $R$
\Ensure Simulation result set $\mathcal{R}$

\State Initialize $\mathcal{R} \leftarrow \emptyset$

\For{each interaction sample $b \in \mathcal{B}$}
    \State Parse $b$ into video $V$, question $Q$, interruption signal $I$, question timestamp $T_0$, and interruption timestamp $T_1$
    
    \Statex \textbf{Step 1: Simulate the initial answering stage}
    \State Extract the observable video segment before the question:
    \State $V_Q \leftarrow V[0:T_0]$
    \State Generate the complete preliminary answer:
    \State $A \leftarrow \mathcal{M}(V_Q, Q)$
    
    \Statex \textbf{Step 2: Simulate interruption timing during generation}
    \State Compute the elapsed speaking time:
    \State $\Delta T \leftarrow \max(0.5, T_1 - T_0)$
    \State Estimate the generated prefix length:
    \State $k \leftarrow \lfloor \Delta T \cdot R \rfloor$
    \State Truncate the preliminary answer to obtain the interrupted partial response:
    \State $A_I \leftarrow \textsc{Truncate}(A, k)$
    
    \Statex \textbf{Step 3: Simulate post-interruption reasoning}
    \State Extract the updated observable video segment up to the interruption time:
    \State $V_I \leftarrow V[0:T_1]$
    \State Construct the interruption prompt $P$ using interruption signal $I$ and task instructions
    \State Query the model with the updated context, original question, and partial response:
    \State $Y \leftarrow \mathcal{M}(V_I, Q, A_I, P)$
    \State Parse $Y$ into predicted interruption type $\hat{c}$ and post-interruption response $A'$
    
    \Statex \textbf{Step 4: Record simulation outputs}
    \State Construct result entry
    \State $r \leftarrow \{Q, I, T_0, T_1, A, A_I, \hat{c}, A'\}$
    \State Add $r$ to $\mathcal{R}$
\EndFor

\State \Return $\mathcal{R}$
\end{algorithmic}
\end{algorithm}

\end{document}